\documentclass{article}
\usepackage{pifont}
\usepackage{inconsolata}
\usepackage{placeins}

\usepackage{times}
\usepackage{microtype}

\usepackage{latexsym}
\usepackage[T1]{fontenc}
\usepackage{inconsolata}
\usepackage[utf8]{inputenc}
\usepackage{microtype}
\usepackage{listings}
\usepackage{url}
\usepackage{xcolor}
\usepackage{algorithmicx}
\usepackage[table]{xcolor}
\definecolor{lightgreen}{rgb}{0.56, 0.93, 0.56}
\usepackage{makecell}
\lstdefinestyle{aclcol}{
  basicstyle=\ttfamily\footnotesize,
  columns=fullflexible,
  keepspaces=true,
  breaklines=true,
  breakatwhitespace=true,
  breakindent=0pt,
  postbreak=\mbox{\textcolor{gray}{$\hookrightarrow$}\space},
  showstringspaces=false
}
\usepackage{inconsolata}
\usepackage{graphicx}
\usepackage{adjustbox}
\usepackage{caption}
\usepackage{algorithm}
\usepackage{array}
\usepackage{tabularx}
\newcolumntype{C}{>{\centering\arraybackslash}X}
\usepackage[noend]{algpseudocode}

\algrenewcommand\alglinenumber[1]{}
\usepackage[table]{xcolor}
\usepackage{booktabs}
\usepackage{pgfplots}
\usepackage{multirow}
\usepackage{adjustbox}
\usepackage{makecell}
\usepackage{graphicx}
\pgfplotsset{compat=1.18}
\usepackage[final]{acl}
\usepackage{listings}
\usepackage{booktabs}
\usepackage{array}
\usepackage{hyperref}
\usepackage{times}
\usepackage{latexsym}

\microtypesetup{protrusion=true,expansion=true}
\usepackage[T1]{fontenc}
\usepackage[utf8]{inputenc}
\usepackage{microtype}
\usepackage{booktabs}
\usepackage{inconsolata}
\usepackage{caption}
\usepackage[most]{tcolorbox}
\usepackage{graphicx}
\usepackage{amsmath}
\usepackage{microtype}
\usepackage{url}
\usepackage{enumitem}
\usepackage[most]{tcolorbox}
\usepackage{tikz}
\usetikzlibrary{arrows.meta,fit,positioning,backgrounds}
\tikzset{
  box/.style={draw, rounded corners, thick, align=center, inner sep=3.5pt, outer sep=2pt, font=\small},
  edge/.style={-{Latex[length=2.5mm]}, very thick},
  edgelite/.style={-{Latex[length=2.0mm]}, thick},
  dashededge/.style={-{Latex[length=2.0mm]}, dashed, thick},
}
\usepackage{tabularx,booktabs,makecell,array}
\newcolumntype{Y}{>{\raggedright\arraybackslash}X}
\DeclareRobustCommand{\method}{De{\scshape alog}}

\title{\method: Decentralized Multi-Agents Log-Mediated Reasoning Framework}

\renewcommand{\arraystretch}{0.85}
\newcommand{\corrauthor}{\thanks{Corresponding author}}
\author{%
  \textbf{Abhijit Chakraborty}\corrauthor\textsuperscript{1} \quad
    \textbf{Ashish Raj Shekhar}\textsuperscript{1}\thanks{These authors contributed equally to this work.} \quad
    \textbf{Shiven Agarwal}\textsuperscript{1}\footnotemark[2] \quad
  \textbf{Vivek Gupta}\footnotemark[1]\textsuperscript{1} \\[4pt]
  \textsuperscript{1}Arizona State University \\[2pt]
  {\tt \{achakr40,vgupt140\}@asu.edu}
}

\begin{document}
\maketitle

\begin{abstract}
Complex multi-hop question answering over text, tables, and images requires integrating heterogeneous evidence with inspectable reasoning. Single-LLM chain-of-thought is opaque, while planner--executor pipelines concentrate faults in a central plan, propagating early errors. We introduce \method, a
decentralized, central-planner-free multi-agent framework whose specialized agents, i.e., Table, Context, Visual, Summarizing, and Verification, coordinate only by reading and appending to a shared, typed natural-language log as persistent memory, with a lightweight scheduler rather than a planner. Shared memory itself is not new: blackboard architectures and recent LLM shared-memory systems all use one, but each retains a \emph{semantic} control unit (a manager, judge, or LLM agent selector). Our contribution is the control design and the finding it isolates: \method's scheduler is \emph{non-semantic}, and under matched agents, retriever, prompts, and corruption seeds, semantic central control is the robustness bottleneck. Rather than competing on raw accuracy, \method{} targets what a central planner forfeits:
robustness, since no single plan concentrates errors and evidence stays globally visible, and transparency, since the append-only log yields a provenance-anchored, human-readable reasoning trace. At 30\% evidence corruption \method{} loses 6.4\% relative EM versus 18--31\% for blackboard and planner systems under the same harness, and swapping in an LLM controller alone degrades $2.5\times$ more.
\end{abstract}

\section{Introduction}
\begin{figure}[htbp]
\centering
\includegraphics[width=0.95\linewidth]{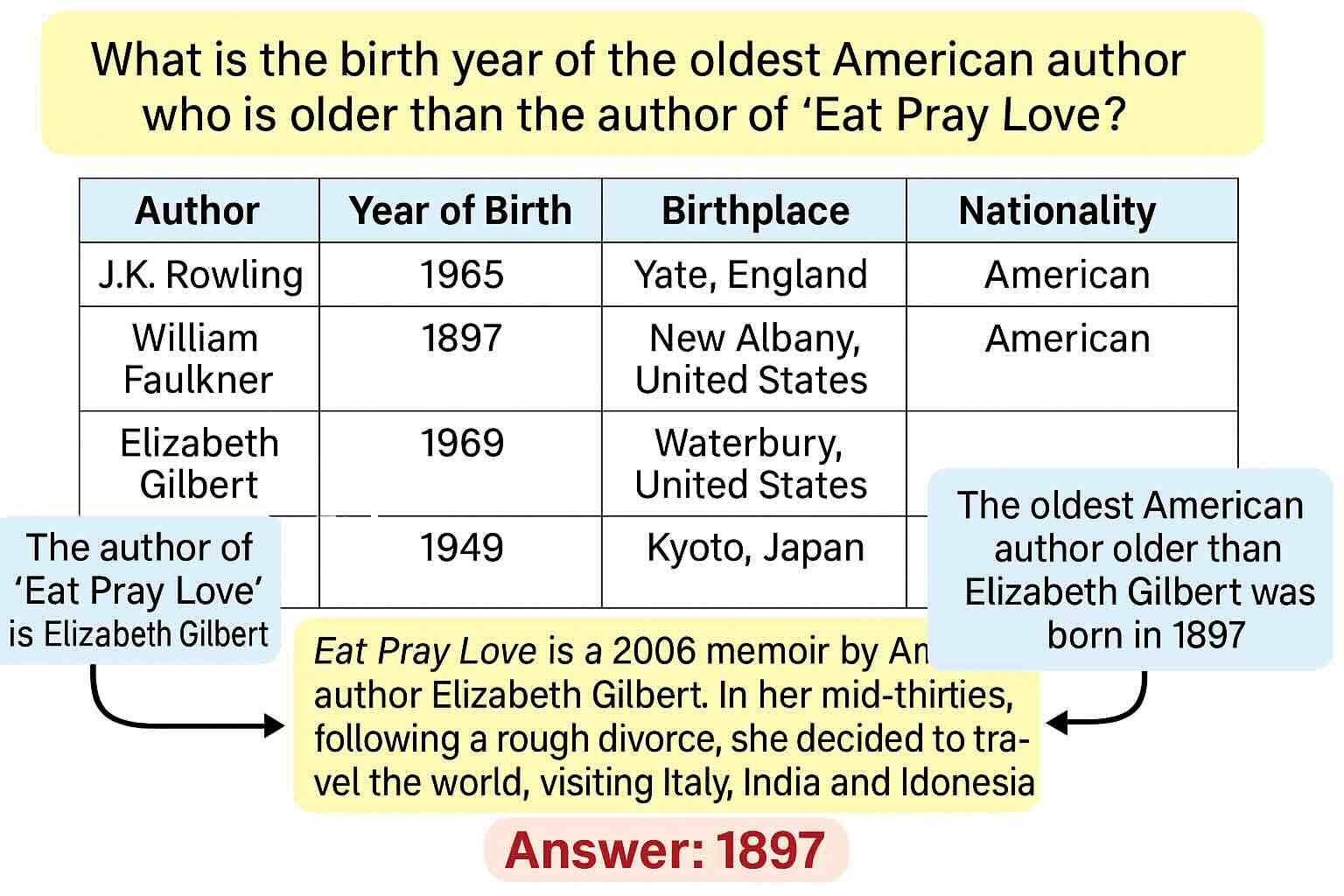}
\caption{\small Example of multi-hop table question answering. The system identifies the author of \emph{``Eat Pray Love''}, extracts metadata from a passage and a table, reasons over birth years and nationalities to find the correct answer.}
\label{fig:example}
\vspace{-1.5em}
\end{figure}

Multi-hop question answering (QA) over complex data, such as tables, textual passages, and images, requires compositional reasoning and the integration of heterogeneous evidence. For example (Fig.~\ref{fig:example}), answering \emph{``What is the birth year of the oldest American author who is older than the author of \textit{Eat Pray Love}?}'' involves identifying entities from a table, extracting attributes from text, and reasoning jointly over nationality and birth years.
To address such multi-hop reasoning, prior work has explored three main directions: (i) \emph{Table QA} approaches based on semantic parsing and neural modeling (e.g., \textsc{WikiTableQuestions} and successors such as \textsc{TAPEX}, \textsc{TAPAS}, \textsc{TaCube}, \textsc{OmniTab}, and \textsc{Lever}) \citep{pasupat-liang-2015-compositional,liu2022tapex,Herzig_2020,yin2022tacube,jiang2022omnitab,ni_lever_2023}; (ii) \emph{reasoning-by-prompting} methods that rely on chain-of-thought or structured prompting (e.g., \textsc{ReAcTable}, Trees/Graphs of Thought) \citep{wei2022chain,zhang2024reactable,besta2024graph}; and (iii) \emph{multi-agent or tool-based pipelines}, typically organized as planner--executor architectures (e.g., \textsc{HuggingGPT}, \textsc{Binder}, \textsc{TabSQLify}, REWoO, \textsc{AutoTQA}, \textsc{MACT}) \citep{shen2023hugginggpt,cheng2023binder,nahid2024tabsqlifyenhancingreasoningcapabilities,xu_rewoo_2023,zhu2024autotqa,zhou2025efficientmultiagentcollaborationtool}. Despite their differences, these approaches largely occupy two ends of a design spectrum. \textbf{Single-LLM methods} based on chain-of-thought prompting~\cite{wei2022chain} perform reasoning implicitly within a single context window; while effective for short or shallow queries, they struggle to maintain explicit intermediate state, verify partial results, and remain stable under long or branching reasoning chains. At the other extreme, \textbf{planner-based pipelines} externalize reasoning through explicit plans and tool invocations, enabling stepwise execution and re-planning based on intermediate outputs. However, this explicit structure introduces a complementary limitation: errors made early in the plan tend to propagate through downstream steps, concentrating uncertainty in the central planner and limiting robustness and transparency. \emph{Together, these limitations call for a framework that exposes intermediate
structure and evidence dependencies without committing to a single, fixed plan that can amplify early errors.} We introduce \method{}, a decentralized, central-planner-free framework in which specialized agents like Table, Context, Visual, Summarizing, and Verification, never communicate directly, but read and append \emph{typed} entries to a shared natural-language log that serves at once as persistent memory and a human-readable reasoning trace. A \textbf{SummarizingAgent} drafts an answer and a \textbf{VerificationAgent} cross-checks it against the log, while a lightweight scheduler manages turn-taking and termination but prescribes no plan.

\noindent\textbf{What is new, and what is not.} A shared memory among specialists is a classical idea: blackboard systems~\citep{ErmanEtAl1980HearsayII,nii_blackboard_1986} and their LLM-era reboots such as \textsc{MetaGPT}~\citep{hong2024metagpt}, the LLM-blackboard of \textsc{LbMAS}~\citep{han_lbmas_2025}, and \textsc{Blackboard-DS}~\citep{salemi2025blackboard} all coordinate through one. We do not claim the substrate. Every prior shared-memory system retains a \emph{semantic} control unit (a manager, judge, or LLM that reads the question and selects the next agent); \method's scheduler is \emph{non-semantic} and never inspects question content, assigns subtasks, or re-plans (Table~\ref{tab:sched_vs_plan}). The contribution is thus an isolated \emph{finding}: with agents, retriever, prompts, and corruption seeds fixed and only the controller swapped, semantic central control is the robustness bottleneck. Blackboard systems under our harness degrade $2.8$--$4.8\times$ more than \method{} at 30\% corruption, and an LLM controller on \method's own agents degrades $2.5\times$ more (\S\ref{sec:mechanism}). Because raw accuracy under a matched retrieval-and-extraction harness offers limited
signal about the role of coordination, we center our claims on the two axes a central planner forfeits. \emph{Robustness:} with no single plan to concentrate errors and all evidence globally visible, corrupted intermediate evidence is outweighed by independently grounded peer
entries rather than propagated downstream, with the verifier contributing a safe fallback that prevents committing a faulty correction. \emph{Transparency:} because every step is recorded as a typed, provenance-anchored entry, the reasoning trace is
provenance-recorded by construction. We characterize both axes through controlled comparisons against a centralized-plan Planner variant and a Plan$\rightarrow$Log hybrid under corrupted inputs and long-horizon reasoning, and against an external planner under corruption, motivating the removal of the central planner in favor of decentralized, log-mediated coordination.
\newline Our framework offers the following key contributions:
\begin{enumerate}[nosep,leftmargin=*]
\item \textbf{Non-semantic control as a design principle:} A decentralized design in which specialized agents communicate only through a typed natural-language log under a deterministic, question-agnostic scheduler with verifier-triggered re-engagement; we
formalize the scheduler--planner boundary across nine dimensions
(Table~\ref{tab:sched_vs_plan}), report the scheduler's measured control-flow statistics (\S\ref{sec:mechanism}), and isolate the controller empirically (Table~\ref{tab:mechanism}).

\item \textbf{Semantic central control is the robustness bottleneck:} A controlled robustness study comparing \method{} against internal centralized-plan variants (Planner, Plan$\rightarrow$Log, Planner++), external planners (AutoTQA, REWoO, HuggingGPT), and shared-memory blackboard systems (\textsc{LbMAS}, \textsc{Blackboard-DS}, \textsc{MetaGPT}) under synthetic, realistic, and \emph{naturally occurring} evidence failures and long-horizon reasoning, with an attribution ablation locating the gain in global evidence visibility rather than the verifier.

\item \textbf{Transparency via a provenance-anchored log:} The typed, append-only log is a provenance-anchored reasoning trace in which each step is tied to its source; we introduce a log-groundedness metric and a counterfactual \emph{necessity} test showing that cited entries are causally used (ablating one flips the answer 60--70\% of the time). (We claim grounding and provenance only, not auditability or a human-usability result.)

\item \textbf{Matched-condition evaluation and ablations:} We evaluate \method{} on six datasets, i.e., FinQA~\cite{chen2021finqa}, TAT-QA~\cite{zhu2021tatqa}, WikiTableQuestions~\cite{pasupat-liang-2015-compositional}, FeTaQA~\cite{nan2021fetaqafreeformtablequestion}, CRT-QA~\cite{zhang_crt-qa_2023}, and
MultiModalQA~\cite{talmor2021multimodalqacomplexquestionanswering}, against 11 baselines on three backbones under a shared retriever and extraction harness, with 95\% bootstrap CIs, and ablate the shared log, agent specialization, verification, deduplication, and context-window sensitivity.
\end{enumerate}

\section{\method ~Framework}\label{sec:method}
\begin{figure}[t]
  \centering
  \includegraphics[
    width=\columnwidth,
    height=.70\textheight,
    keepaspectratio,
    trim=8pt 10pt 8pt 10pt,clip
  ]{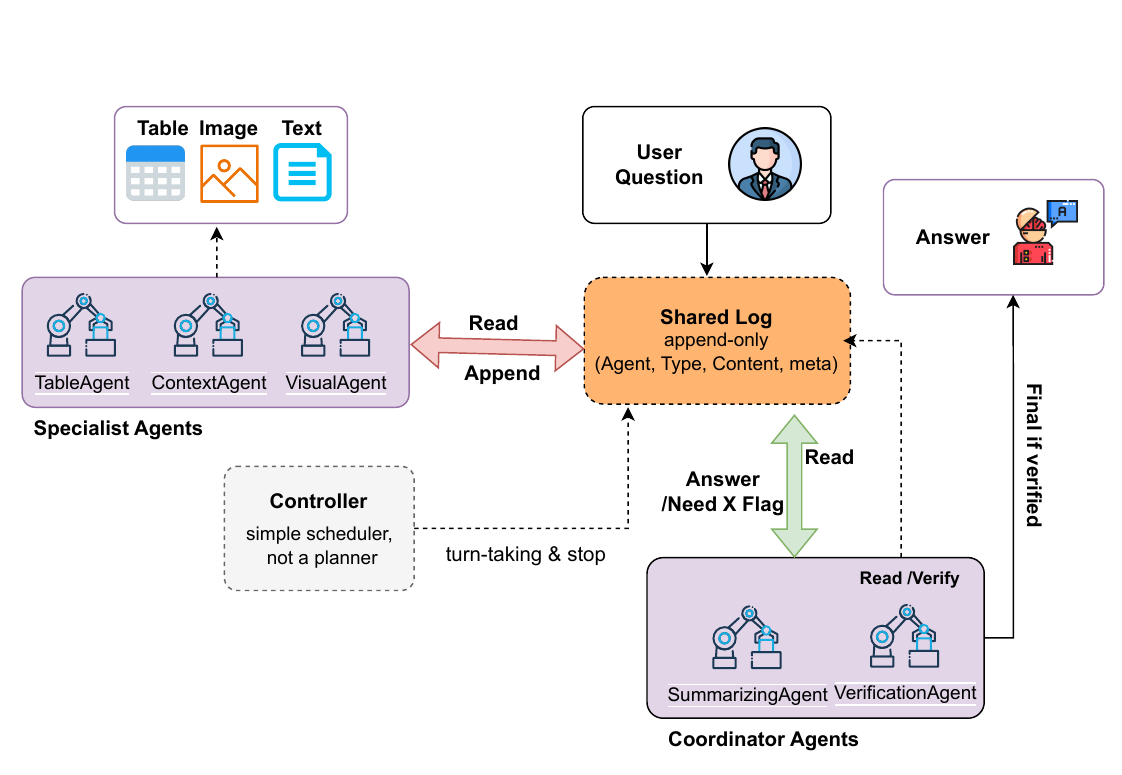}
  \caption{\small\method: central-planner-free, log-mediated QA. Specialized agents read from and append typed entries to a shared append-only log; the Summarizer synthesizes a candidate answer and the Verifier cross-checks it. The scheduler is \emph{non-semantic}: it cycles agents in fixed order (Table $\to$ Context $\to$ Visual $\to$ Summarizing $\to$ Verification), gating each turn by the agent's \texttt{should\_act} precondition rather than by task content, and halts on a Verifier \textsc{OK} or at the round budget $R$. A Verifier \textsc{Flag} triggers exactly one re-engagement round in which upstream agents re-read the log and may append corrections.}
  \label{fig:arch}
  \vspace{-1.0em}
\end{figure}

Our approach draws from decentralized team architectures and \emph{shared memory} among agents. Traditional \emph{blackboard systems} enabled expert modules to solve problems via a common blackboard \citep{ErmanEtAl1980HearsayII,nii_blackboard_1986,EngelmoreMorgan1988,Corkill1991}, but always under a \emph{control component} that inspected the blackboard's content to pick the next knowledge source; their LLM-era successors keep that role as an LLM controller \citep{han_lbmas_2025,salemi2025blackboard} or a role/SOP orchestrator \citep{hong2024metagpt}. \emph{Generative agents} have used shared memory for simulations \citep{park2023generativeagentsinteractivesimulacra}. While \textsc{Graph-of-Thoughts} and \textsc{AMAR} explored multi-agent reasoning, they either integrate reasoning within one model or distribute tasks without communication \citep{besta2024graph,sami_adaptive_2025}. This work establishes a \emph{log-mediated} multi-agent reasoning framework for factual QA with five LLM agents: \emph{Table}, \emph{Context}, \emph{Visual}, \emph{Summarizing}, and \emph{Verification}, communicating through a shared log (Fig.~\ref{fig:arch}), and removes the semantic control component entirely (App.~\ref{app:differentiation} tabulates the differences from prior shared-memory systems). Details are in App.~\ref{app:log}.
\newline\textbf{Design goals and assumptions.} Our design emphasizes (i)~\emph{modularity}, enabling plug-in specialists without retraining; (ii)~\emph{transparency}, via a provenance-aware shared log as persistent memory; and (iii)~\emph{robustness}, by keeping early errors from concentrating in a central plan and propagating---with the verifier as a safe fallback rather than an active repair mechanism. We assume access to tabular corpora, text, and images; agents operate zero-/few-shot. Instead of a central planner, a lightweight \emph{scheduler} manages turn-taking, guardrails, and termination without prescribing plans. Its fixed agent order is question-agnostic (a structural prior, not a task plan), so \method{} is \emph{central-planner-free} rather than free of all ordering.
\newline\textbf{Scheduler vs.\ planner: a formal distinction.} A key design choice is replacing a central planner with a lightweight scheduler. Table~\ref{tab:sched_vs_plan} formalizes this across nine dimensions. The \method{} controller operates at the \emph{resource-management} level with no semantic awareness, like an OS process scheduler, while central planners~\citep{shen2023hugginggpt,zhu2024autotqa,xu_rewoo_2023} operate at the \emph{reasoning-control} level. Empirically, under 30\% corruption the planner drops 24.7\% EM and \method{} 6.4\% (Table~\ref{tab:robustness_em}). The learned gate is a binary continue/stop classifier, not a planning component.

\begin{table}[t]
\centering
\small
\setlength{\tabcolsep}{2pt}
\begin{tabular}{@{}p{2.4cm}cc@{}}
\toprule
\textbf{Capability} & \textbf{\method{} Sched.} & \textbf{Central Planner} \\
\midrule
Inspects question semantics & \ding{55} & \ding{51} \\
Generates task graphs & \ding{55} & \ding{51} \\
Assigns subtasks & \ding{55} & \ding{51} \\
Re-plans on failure & \ding{55} & \ding{51} \\
Controls reasoning strategy & \ding{55} & \ding{51} \\
Semantic awareness & None & Full \\
Single point of failure & \ding{55} & \ding{51} \\
Learned gating role & Stop/cont. & Absent \\
Analogy & OS sched. & Project mgr \\
\bottomrule
\end{tabular}
\vspace{-0.5em}
\caption{\small Scheduler vs.\ planner distinction. The scheduler manages \emph{when} agents run, not \emph{what} they do.}
\label{tab:sched_vs_plan}
\vspace{-1.5em}
\end{table}

\vspace{-0.5em}
\paragraph{Agent roles and I/O contracts.} The \textbf{TableAgent} parses tables and posts cell-level facts. The \textbf{ContextAgent} retrieves passages. The \textbf{VisualAgent} converts charts to text. The \textbf{SummarizingAgent} composes answers or summaries. The \textbf{VerificationAgent} checks calculations and factual support. Each agent implements \texttt{should\_act(log)} (trigger heuristic) and \texttt{act(log)} (LLM call for typed entry). The \textbf{VerificationAgent} checks consistency, recalculates, verifies units, and ensures answers are supported by log entries. On detecting inconsistencies, it re-engages. Unlike frameworks using triples or code, the verifier operates on unstructured logs without curated error taxonomies. \method's faithfulness stems from multi-agent consensus grounded in the shared log.
\newline\textbf{Shared log and types.} Each entry is a tuple $(\texttt{Agent},\texttt{Type},\texttt{Content},\texttt{meta})$ where \texttt{meta} stores step index, time, and provenance. We use a fixed vocabulary of \texttt{Type}: \textsc{Lookup}, \textsc{Quote}, \textsc{Visual}, \textsc{Summary}, \textsc{Answer}, and \textsc{Flag}/\textsc{OK}. All agents have \emph{global visibility}; ablations show restricting visibility markedly hurts performance. We deduplicate near-duplicate entries using ROUGE-L similarity (threshold~$>0.85$) combined with per-agent caps, saving 13.3\% of tokens with negligible accuracy impact ($\Delta$\,EM\,$=-0.008$; Table~\ref{tab:dedup_ablation}). Pseudocode for the controller loop is given in Algorithm~\ref{alg:logqa_moved} (App.~\ref{app:moved_tables}).
\newline\textbf{Controller loop.} Each round, the scheduler invites the Table, Context, and Visual agents to contribute; when new evidence appears, the Summarizing Agent drafts an answer, which the Verification Agent checks before stopping. A \textsc{Flag} triggers one re-engagement round to fix the issue. Most questions resolve in $3$--$4$ agent calls; we cap rounds at $R\!=\!6$ and apply per-agent and duplicate-entry limits as safety guardrails.
\newline\textbf{Gating.} A small learned gate decides when to stop retrieving. It is trained on WikiTQ traces disjoint from the evaluation split, then applied unchanged to all six datasets while saving $9$--$18\%$ of tokens on tabular benchmarks with negligible accuracy cost, and (correctly) never firing on CRT-QA, where longer reasoning matters (Table~\ref{tab:efficiency_gating}). Crucially, \emph{headline accuracy in Tables~\ref{tab:FeTaQA_FinQA_TATQA} and \ref{tab:MMQA_WikiTQ} is reported without the gate}, so the accuracy claim cannot depend on the classifier's quality.
\newline\textbf{Memory.} Older log entries are summarized with citations to fit the context window; doubling the window from $4$k to $8$k tokens yields only $+0.013$ EM on average (Table~\ref{tab:context_ablation}), confirming that reasoning difficulty, not context length, is the main bottleneck.

We provide full prompts, log types, regex patterns, no-progress criteria, and truncation rules in App.~\ref{app:log} and the public code release.\footnote{\url{https://github.com/CoRAL-ASU/Agentic_ADAF.git}} The repository provides the complete code, prompts, corruption generators, shared seeds, and per-example noise keys needed to reproduce Table~\ref{tab:robustness_em}, together with representative log traces for audit.

\section{Experiments}
\label{sec:experiments}
\par\indent\textbf{Setup:} We evaluate \method~and baselines on six multi-hop question answering datasets: FinQA, TAT-QA, WikiTableQuestions (WikiTQ), FeTaQA, CRT-QA, and MMQA. These datasets cover numerical reasoning, compositional queries, and open-domain retrieval. All datasets use matched backbone configurations and identical test sets, adhering to default splits and evaluation protocols. Dataset-specific statistics, such as the number of questions and modality composition, are in Table~\ref{tab:dataset_stats} (App.~\ref{app:dataset_stats}).
\newline\emph{Input Filtering and Retrieval:} We use a consistent BM25+miniLM~\cite{robertson_probabilistic_2009,wang_minilm_2020} retriever for input filtering and retrieval across all methods. This retriever extracts relevant passages, table segments, or image captions as input context. No task-specific filtering or preprocessing is applied to maintain comparability.
\newline\emph{Metrics and Normalization.} Table~\ref{tab:metrics_protocol} (App.~\ref{app:metrics}) specifies the exact metric per dataset (EM, or the official scorer; TAT-QA reports $(\text{EM}+\text{F1})/2$; FeTaQA EM* applies a uniform \texttt{Answer:}-suffix extraction rule to every method). All normalization rules (numeric parsing, unit canonicalization, floating-point tolerance $\pm 0.01$) are implemented in a shared function used identically across all methods.
\paragraph{Per-dataset evaluation protocol.}Every dataset uses its official scorer on its default test split. The pipeline is closed-book i.e. no Wikipedia, web search, or knowledge graphs. Retrieval only filters the evidence shipped with each example, never an external corpus. Every method gets the same evidence pool and the same scorer (see Tables~\ref{tab:FeTaQA_FinQA_TATQA} and~\ref{tab:MMQA_WikiTQ}).
\newline\textbf{Evaluation Approach:} Accuracy is the primary metric, computed either as exact match or via each dataset's official scoring rules (Table~\ref{tab:metrics_protocol}). We define a catastrophic error (Figure~\ref{fig:catastrophic_err}) as a prediction that is empty, off by more than 50\% on a numeric question, or that mentions entities not present in the evidence, these are the failures we most want to avoid. An operator chain (Tables~\ref{tab:robustness_em}, \ref{tab:long_horizon}) measures the minimum number of retrieval, arithmetic, comparison, or lookup steps needed to reach the correct answer; we read this from the released reasoning programs where available, and count distinct evidence hops otherwise. All results are averaged over five random trials, reported with 95\% bootstrap confidence intervals (1{,}000 resamples)~\cite{zrimsek_quantifying_2024} in Tables~\ref{tab:FeTaQA_FinQA_TATQA} and \ref{tab:MMQA_WikiTQ}. Where we report a per-column best, we treat overlapping 95\% CIs over the shared seeds as statistically indistinguishable rather than as a win. Efficiency is reported as LLM calls per query, total tokens used, and per-query latency (mean and P90). Corruption degradation is reported as a \emph{relative} drop, $(\text{clean}-\text{corrupted})/\text{clean}$, except in the attribution ablation (Table~\ref{tab:mechanism}a), which reports absolute EM points on the dev split.
\newline\textbf{LLM Models:} We test three open backbones (LLaMA-3 8B, Mistral 7B, Qwen-3 8B) on identical footing. Summarizing and Verification agents run at temperature 0; Table and Context agents at 0.3. BLIP-2~\cite{li_blip-2_2023} and PaddleOCR~\cite{cui_paddleocr-vl_2025} process images inline, so reported latencies (Tables~\ref{tab:efficiency_partial_fixed}, \ref{tab:stage_latency}, \ref{tab:pareto}) include this cost. Models run on A100s via vLLM.
\newline\textbf{Tool fairness.} Tables~\ref{tab:FeTaQA_FinQA_TATQA} and \ref{tab:MMQA_WikiTQ} use a \emph{tool-restricted} setting: every method receives identical multimodal tools (BLIP-2, PaddleOCR) but no Python interpreter or calculator, isolating reasoning quality from code-execution engineering. Natively code-executing methods (ReAcTable, Codex, Program-of-Thoughts) are additionally evaluated with their interpreter enabled in App.~\ref{app:tool_enabled}; \method{} runs without code execution in both settings.
\newline\textbf{Baseline Methods:}
We compare our proposed approach against a comprehensive set of baselines including multi-hop reasoning methods such as Chain-of-Thought prompting~\cite{wei2022chain}, REWoO~\cite{xu_rewoo_2023}, Chameleon~\cite{lu2023chameleon}, FireAct~\cite{chen_fireact_2023}, Lumos~\cite{yin2023lumos}, and HUSKY~\cite{lin2024husky}; table question answering models like Codex~\cite{chen_evaluating_2021} and TableCritic~\cite{yu_table-critic_2025}; and multi-agent systems including AutoTQA~\cite{zhu2024autotqa}, ReAcTable~\cite{zhang2024reactable} and Dater~\cite{ye2023dater}. We additionally compare against Chain-of-Table~\cite{wang2024chain}, Program-of-Thoughts~\cite{chen2023pot}, and TabSQLify~\cite{nahid2024tabsqlifyenhancingreasoningcapabilities} on the datasets where their public implementations apply (discussed in \S\ref{sec:experiments}). All share comparable LLM backbones and run zero-shot under the same harness.

\subsection{Results and Analysis}
\par\textbf{Accuracy is equalized under matched conditions.} Under a shared BM25+miniLM retriever, a uniform \texttt{Answer:}-suffix extraction rule, and no code execution, agent and prompting strategies cluster tightly: across Tables~\ref{tab:FeTaQA_FinQA_TATQA} and~\ref{tab:MMQA_WikiTQ}, most baselines fall within a few EM points of one another on each dataset, and \method{} is competitive rather than categorically stronger. On FinQA \method{} attains 76.5\% (LLaMA-3 8B) and 76.4\% (Qwen-3 8B), and on FeTaQA 75.8\% and 76.1\%, ahead of multi-agent baselines such as AutoTQA, ReAcTable, and TableCritic but by margins that frequently
overlap at 95\% CIs. The pattern is not uniform: TableCritic leads TAT-QA across all three backbones, and CoT spikes on Mistral 7B. We read this clustering as the central methodological point, i.e., \emph{when retrieval and answer extraction are fixed and code execution is disallowed, raw accuracy reflects the shared harness more than the coordination strategy, and is a weak axis for comparing agent architectures}. We therefore treat accuracy as a competitiveness check and locate \method's contribution in robustness and transparency, the two axes the harness does not equalize.
\begin{table*}[ht]
\small
\centering
\setlength{\tabcolsep}{4pt}
\renewcommand{\arraystretch}{1.05}
\small
\begin{tabular}{lccc}
\toprule
\textbf{Method} & \textbf{FeTaQA} & \textbf{FinQA} & \textbf{TAT-QA} \\
\midrule
CoT         & 54.9$_{\pm 1.9}$ / \textbf{79.7$_{\pm 1.3}$} / 54.6$_{\pm 1.9}$ & 56.8$_{\pm 1.9}$ / \textbf{79.4$_{\pm 1.3}$} / 61.7$_{\pm 1.7}$ & 54.5$_{\pm 1.9}$ / 61.2$_{\pm 1.7}$ / 55.6$_{\pm 1.9}$ \\
REWoO       & 68.5$_{\pm 1.5}$ / 70.4$_{\pm 1.5}$ / 69.7$_{\pm 1.5}$ & 70.0$_{\pm 1.5}$ / 70.3$_{\pm 1.5}$ / 69.7$_{\pm 1.5}$ & 68.5$_{\pm 1.5}$ / 69.7$_{\pm 1.5}$ / 70.0$_{\pm 1.5}$ \\
Chameleon   & 70.2$_{\pm 1.5}$ / 71.8$_{\pm 1.4}$ / 70.5$_{\pm 1.4}$ & 70.7$_{\pm 1.5}$ / 71.9$_{\pm 1.4}$ / 70.1$_{\pm 1.4}$ & 70.2$_{\pm 1.5}$ / 71.4$_{\pm 1.4}$ / 70.1$_{\pm 1.4}$ \\
FireAct     & 72.3$_{\pm 1.4}$ / 72.4$_{\pm 1.3}$ / 72.1$_{\pm 1.4}$ & 72.7$_{\pm 1.4}$ / 72.8$_{\pm 1.3}$ / 72.4$_{\pm 1.4}$ & 72.3$_{\pm 1.4}$ / 72.2$_{\pm 1.3}$ / 72.4$_{\pm 1.4}$ \\
Lumos       & 73.3$_{\pm 1.4}$ / 73.2$_{\pm 1.3}$ / 73.9$_{\pm 1.3}$ & 73.2$_{\pm 1.4}$ / 73.0$_{\pm 1.3}$ / 73.2$_{\pm 1.3}$ & 73.0$_{\pm 1.4}$ / 73.9$_{\pm 1.3}$ / 73.8$_{\pm 1.3}$ \\
HUSKY       & 73.3$_{\pm 1.3}$ / 74.0$_{\pm 1.3}$ / 73.8$_{\pm 1.3}$ & 73.4$_{\pm 1.3}$ / 73.1$_{\pm 1.3}$ / 73.9$_{\pm 1.3}$ & 73.6$_{\pm 1.3}$ / 73.2$_{\pm 1.3}$ / 74.0$_{\pm 1.3}$ \\
AutoTQA     & 69.4$_{\pm 1.5}$ / 70.2$_{\pm 1.5}$ / 69.5$_{\pm 1.5}$ & 69.8$_{\pm 1.5}$ / 70.1$_{\pm 1.5}$ / 69.4$_{\pm 1.5}$ & 69.3$_{\pm 1.5}$ / 69.9$_{\pm 1.5}$ / 70.0$_{\pm 1.5}$ \\
Dater       & 68.5$_{\pm 1.6}$ / 69.2$_{\pm 1.5}$ / 68.7$_{\pm 1.6}$ & 68.9$_{\pm 1.6}$ / 69.3$_{\pm 1.5}$ / 68.2$_{\pm 1.6}$ & 68.7$_{\pm 1.6}$ / 68.1$_{\pm 1.5}$ / 68.8$_{\pm 1.6}$ \\
ReAcTable   & 69.0$_{\pm 1.5}$ / 70.2$_{\pm 1.4}$ / 72.7$_{\pm 1.3}$ & 69.4$_{\pm 1.5}$ / 70.1$_{\pm 1.4}$ / 72.8$_{\pm 1.3}$ & 69.2$_{\pm 1.5}$ / 69.5$_{\pm 1.4}$ / 74.3$_{\pm 1.3}$ \\
Codex       & 68.5$_{\pm 1.6}$ / 69.4$_{\pm 1.5}$ / 74.0$_{\pm 1.3}$ & 68.2$_{\pm 1.6}$ / 69.1$_{\pm 1.5}$ / 73.3$_{\pm 1.3}$ & 68.1$_{\pm 1.6}$ / 68.7$_{\pm 1.5}$ / 73.3$_{\pm 1.3}$ \\
TableCritic & 73.1$_{\pm 1.3}$ / 74.2$_{\pm 1.3}$ / 75.2$_{\pm 1.2}$ & 73.4$_{\pm 1.3}$ / 74.5$_{\pm 1.3}$ / 75.2$_{\pm 1.2}$ & \textbf{75.4$_{\pm 1.3}$} / \textbf{75.7$_{\pm 1.3}$} / \textbf{74.8$_{\pm 1.2}$} \\
\rowcolor{lightgreen}
\method     & \textbf{75.8$_{\pm 1.1}$} / 76.9$_{\pm 1.1}$ / \textbf{76.1$_{\pm 1.0}$} & \textbf{76.5$_{\pm 1.1}$} / 75.2$_{\pm 1.1}$ / \textbf{76.4$_{\pm 1.0}$} & 72.4$_{\pm 1.2}$ / 73.1$_{\pm 1.2}$ / 71.2$_{\pm 1.2}$ \\
\bottomrule
\end{tabular}
\caption{\small Results (\%) on FeTaQA (EM*), FinQA (EM), TAT-QA (mean of the official EM and F1, $(\text{EM}+\text{F1})/2$; Table~\ref{tab:metrics_protocol}) with $\pm$95\% bootstrap CIs (1{,}000 resamples, 5 seeds). Format: LLaMA-3 8B / Mistral 7B / Qwen-3 8B. Bold = small differences within overlapping CIs.}
\label{tab:FeTaQA_FinQA_TATQA}
\end{table*}

\begin{table*}[ht]
\centering
\setlength{\tabcolsep}{4pt}
\renewcommand{\arraystretch}{1.05}
\small
\begin{tabular}{lccc}
\toprule
\textbf{Method} & \begin{tabular}[c]{@{}c@{}}\textbf{MMQA} \\ \textbf{(full)}\end{tabular} & \begin{tabular}[c]{@{}c@{}}\textbf{MMQA} \\ \textbf{(text/table)}\end{tabular} & \begin{tabular}[c]{@{}c@{}}\textbf{WikiTQ}\end{tabular} \\
\midrule
CoT         & 56.4$_{\pm 1.9}$ / 58.2$_{\pm 1.3}$ / 68.9$_{\pm 1.6}$ & 56.5$_{\pm 1.8}$ / 55.2$_{\pm 1.2}$ / 56.1$_{\pm 1.8}$ & 53.1$_{\pm 1.9}$ / 58.7$_{\pm 1.3}$ / 64.4$_{\pm 1.6}$ \\
REWoO       & 70.0$_{\pm 1.5}$ / 67.8$_{\pm 1.5}$ / 69.5$_{\pm 1.5}$ & 69.9$_{\pm 1.5}$ / 69.6$_{\pm 1.5}$ / 70.4$_{\pm 1.5}$ & 68.6$_{\pm 1.5}$ / 70.5$_{\pm 1.5}$ / 71.0$_{\pm 1.5}$ \\
Chameleon   & 70.4$_{\pm 1.5}$ / 72.5$_{\pm 1.4}$ / 70.9$_{\pm 1.4}$ & 71.4$_{\pm 1.4}$ / 72.7$_{\pm 1.4}$ / 70.3$_{\pm 1.4}$ & 69.4$_{\pm 1.5}$ / 71.8$_{\pm 1.4}$ / 70.1$_{\pm 1.4}$ \\
FireAct     & 72.7$_{\pm 1.4}$ / 72.8$_{\pm 1.3}$ / 71.2$_{\pm 1.4}$ & 72.5$_{\pm 1.4}$ / 73.5$_{\pm 1.3}$ / 71.8$_{\pm 1.4}$ & 71.0$_{\pm 1.4}$ / 72.4$_{\pm 1.3}$ / 71.9$_{\pm 1.4}$ \\
Lumos       & 73.6$_{\pm 1.4}$ / 73.4$_{\pm 1.3}$ / 73.7$_{\pm 1.3}$ & 72.5$_{\pm 1.3}$ / 73.1$_{\pm 1.3}$ / 72.5$_{\pm 1.3}$ & 71.1$_{\pm 1.4}$ / 74.7$_{\pm 1.3}$ / 73.7$_{\pm 1.3}$ \\
HUSKY       & 73.6$_{\pm 1.3}$ / 74.9$_{\pm 1.3}$ / 73.6$_{\pm 1.3}$ & 72.8$_{\pm 1.3}$ / 73.2$_{\pm 1.3}$ / 73.7$_{\pm 1.3}$ & 72.8$_{\pm 1.3}$ / 73.5$_{\pm 1.3}$ / 74.0$_{\pm 1.3}$ \\
AutoTQA     & 69.7$_{\pm 1.5}$ / 70.4$_{\pm 1.5}$ / 69.0$_{\pm 1.5}$ & 69.7$_{\pm 1.5}$ / 70.3$_{\pm 1.5}$ / 68.8$_{\pm 1.5}$ & 68.0$_{\pm 1.5}$ / 71.9$_{\pm 1.4}$ / 70.0$_{\pm 1.5}$ \\
Dater       & 68.9$_{\pm 1.6}$ / 69.3$_{\pm 1.5}$ / 68.4$_{\pm 1.6}$ & 68.7$_{\pm 1.6}$ / 69.7$_{\pm 1.5}$ / 67.4$_{\pm 1.6}$ & 68.1$_{\pm 1.6}$ / 69.2$_{\pm 1.5}$ / 68.7$_{\pm 1.6}$ \\
ReAcTable   & 69.0$_{\pm 1.5}$ / 70.3$_{\pm 1.4}$ / 73.4$_{\pm 1.3}$ & 68.7$_{\pm 1.5}$ / 70.2$_{\pm 1.4}$ / 73.6$_{\pm 1.3}$ & 68.5$_{\pm 1.5}$ / 70.9$_{\pm 1.4}$ / 72.5$_{\pm 1.3}$ \\
Codex       & 69.0$_{\pm 1.6}$ / 69.9$_{\pm 1.5}$ / 74.9$_{\pm 1.3}$ & 67.4$_{\pm 1.6}$ / 69.1$_{\pm 1.5}$ / 73.6$_{\pm 1.3}$ & 66.4$_{\pm 1.6}$ / 70.3$_{\pm 1.5}$ / 73.5$_{\pm 1.3}$ \\
TableCritic & 73.2$_{\pm 1.3}$ / 73.4$_{\pm 1.3}$ / 75.4$_{\pm 1.2}$ & 73.8$_{\pm 1.3}$ / 74.0$_{\pm 1.3}$ / 75.4$_{\pm 1.2}$ & 72.7$_{\pm 1.3}$ / 74.4$_{\pm 1.3}$ / \textbf{76.6$_{\pm 1.2}$} \\
\rowcolor{lightgreen}
\method     & \textbf{76.7$_{\pm 1.0}$} / \textbf{76.8$_{\pm 1.1}$} / \textbf{76.2$_{\pm 1.0}$} & \textbf{74.2$_{\pm 1.1}$} / \textbf{76.0$_{\pm 1.1}$} / \textbf{75.5$_{\pm 1.0}$} & \textbf{75.2$_{\pm 1.1}$} / \textbf{76.5$_{\pm 1.1}$} / 75.3$_{\pm 1.0}$ \\
\bottomrule
\end{tabular}
\caption{\small Accuracy (\%) on MMQA full, MMQA text/table, and WikiTQ with $\pm$95\% bootstrap CIs (1{,}000 resamples, 5 seeds). Format: LLaMA-3 8B / Mistral 7B / Qwen-3 8B. Bold = small differences within overlapping CIs. \method{} attains the per-column best on every column except WikiTQ / Qwen-3, where TableCritic reports 76.6\% (CI overlaps with \method's 75.3\%).}
\label{tab:MMQA_WikiTQ}
\end{table*}

These comparisons confirm that, under matched conditions, no coordination strategy dominates on accuracy; specialized systems and \method{} alike sit within a narrow band, including on MMQA and WikiTQ (Table~\ref{tab:MMQA_WikiTQ}); we make no claim of a clean-accuracy advantage. This makes accuracy \emph{alone} insufficient, rather than uninformative, for our central question of what decentralization buys: a few cells still separate (e.g.\ TableCritic on TAT-QA, \method{} on MMQA/WikiTQ), but the spreads are narrow under matched conditions, which motivates the robustness and transparency analyses that follow.

\paragraph{Tool-enabled comparison.}
\method{} runs without a code interpreter in both settings of Tables~\ref{tab:tool_enabled} and~\ref{tab:tool_enabled_mmqa} (App.~\ref{app:tool_enabled}; PoT rationale in \S\ref{sec:experiments}). Tool gains on FinQA average $+2.0$ EM across baselines, yet \method{} without code execution still wins on 2 of 3 backbones and is within $0.4$ EM on the third; and on MMQA and WikiTQ (Table~\ref{tab:tool_enabled_mmqa}, dev split) \method{} without tools exceeds even the \emph{tool-enabled} baselines on all three backbones, indicating its advantage there is not attributable to tool access.
\paragraph{Efficiency and Resource Utilization:}
\method{} uses fewer LLM calls than most multi-agent baselines ($3.1$ vs.\ $3.4$--$3.8$; Table~\ref{tab:efficiency_partial_fixed}) but the highest \emph{sequential} latency ($0.95$s); the parallel variant Par-R6 cuts this to $0.68$s at no accuracy cost and is the production operating point (Pareto sweep in Table~\ref{tab:pareto}). Learned gating reduces tokens by up to 18\% and latency by up to $2.18\times$ at $\le0.01$ EM cost (Table~\ref{tab:efficiency_gating}), because the gate fires preferentially on the most expensive examples.

\paragraph{Ablations and log behavior.} Doubling the context window (4k$\to$8k) yields only $+0.013$ average EM (Table~\ref{tab:context_ablation}, App.~\ref{app:moved_tables})which is a reasoning difficulty, not context length, is the bottleneck. Long traces degrade gracefully: accuracy stays high through 5--6 steps and $0.88$ at 7--8 on MMQA-dev slices (sample-size caveat, Table~\ref{tab:long_horizon}), settling at $\sim$0.70 on the larger CRT-QA test set. Removing ROUGE-L deduplication adds $13.3\%$ tokens at $\Delta$EM$=0.000$ (Table~\ref{tab:dedup_ablation}; per-agent rates in Table~\ref{tab:per_agent_dup}, App.~\ref{app:moved_tables}); with gating it gives $\sim$25\% token reduction. Per-method latency is in Figure~\ref{fig:latency_comparison} (App.~\ref{app:moved_tables}).

\vspace{-0.25em}
\paragraph{Corruption protocol.}
For the robustness experiments in Tables~\ref{tab:robustness_em} and~\ref{tab:fault_injection}, we inject controlled corruption into both (i)~retrieved text spans returned by the BM25+miniLM retriever and (ii)~table cell values from the source tables, under two families. \emph{Structural/numeric} noise samples one of three operations with equal probability per corrupted item: \emph{numeric perturbation} (replace a numeric cell or quoted number with a value drawn uniformly from a $\pm 10\%$ band around the original), \emph{row swap} (exchange the contents of two random rows within the same table), and \emph{deletion} (remove the span or replace the cell with an empty token). \emph{Semantic} noise samples between \emph{paraphrase substitution} (replace a span with a fluent but content-altering LLM paraphrase) and \emph{entity swap} (replace a named entity or quantity with a plausible same-type distractor drawn from the example's own evidence). Corruption is applied at two stages: \emph{before any agent reads the evidence} (so retrieval and table-parsing agents see corrupted inputs), and additionally to log entries posted by \texttt{TableAgent} and \texttt{ContextAgent} before downstream agents consume them (the same in-log injection point used for the fault-injection ablation in Table~\ref{tab:fault_injection}). The corruption \emph{rate} (0\%, 10\%, 20\%, 30\% in Table~\ref{tab:robustness_em}; 10\%, 20\%, 30\% in Table~\ref{tab:fault_injection}) is the fraction of evidence items perturbed in a given run. We run the protocol on MMQA and, as a second table-centric dataset, TAT-QA. The corruption draws are keyed by (seed, example) and are therefore shared across all nine systems, i.e., Planner, Plan$\rightarrow$Log, Planner++, AutoTQA, REWoO, the three shared-memory systems \textsc{LbMAS}, \textsc{Blackboard-DS}, and \textsc{MetaGPT}, and \method{} (Planner++ and the shared-memory systems on the panels where they are run), so every column of Table~\ref{tab:robustness_em} is evaluated on identical corrupted inputs (paired comparisons over five seeds). \emph{Calibration.} The synthetic rates are not arbitrary: on the clean MMQA pipeline we measured the organic frequency of each upstream failure, i.e., retrieval miss $14\%$, OCR misread $13\%$, caption error $10\%$, and table-parse error $9\%$ (App.~\ref{app:error_distribution}), so the $10$--$30\%$ band brackets what the deployed pipeline actually produces.

\paragraph{Planner and Plan$\rightarrow$Log: controlled architectural variants.}
The Planner and Plan$\rightarrow$Log comparators (Table~\ref{tab:robustness_em}, Table~\ref{tab:fault_injection}, Figure~\ref{fig:catastrophic_err}) are deliberately \emph{minimal internal variants} of \method{} that isolate one variable: whether the system commits to a centralized natural-language plan before reasoning begins. They are not SoTA external planners, those (REWoO, AutoTQA, Chameleon, Lumos, FireAct) are evaluated on clean inputs in Tables~\ref{tab:FeTaQA_FinQA_TATQA}--\ref{tab:MMQA_WikiTQ}. All three share \method's backbone, retriever, prompt budget, and guardrails; only the planning component varies.

\emph{Planner} is a single LLM call that plans, solves, and answers; \emph{Plan$\rightarrow$Log} seeds a heuristic plan into the normal \method{} coordinator and never revises it, so the gap to \method{} measures the \emph{cost of pre-committing to a plan under noise} with LLM, retriever, and stopping fixed (pseudocode and details in App.~\ref{app:control}).

\paragraph{External validity under real planners.}
Two external multi-agent planners, AutoTQA~\cite{zhu2024autotqa} and REWoO~\cite{xu_rewoo_2023}, run under the identical protocol and seeds degrade far more than \method{} under structural noise ($21.4\%$ and $21.1\%$ relative vs.\ $6.4\%$), tracking the internal Planner ($24.7\%$), and the ordering holds under semantic noise ($13.6\%$ vs.\ $19.2\%$). The internal Planner is thus not a strawman: real planners share its robustness class regardless of sophistication. A third planner, HuggingGPT~\citep{shen2023hugginggpt}, on HybridQA reproduces the ordering (App.~\ref{app:extra_planner_dataset}).
\paragraph{Robustness to evidence corruption.} \method's advantage is on robustness, not clean accuracy. Under structural MMQA noise, \method{} retains $73.0\%$ EM at $30\%$ corruption, versus $55.0$ (internal Planner), $66.0$ (Plan$\rightarrow$Log), $64.0$ (the budget-matched Planner++), $55.0$ (AutoTQA), and $52.7$ (REWoO)---a $6.4\%$ relative drop against $24.7\%$ (Planner), $21.4\%$ (AutoTQA), and $21.1\%$ (REWoO) (Table~\ref{tab:robustness_em}, Figure~\ref{fig:catastrophic_err}); the gap to AutoTQA widens from $8.0$ points on clean inputs to $18.0$ points at $30\%$. The pattern persists under \emph{semantic} noise ($67.4\%$ vs.\ $61.4\%$ for the strongest comparator at $30\%$) and on the second dataset, TAT-QA (Table~\ref{tab:robustness_tatqa}, App.~\ref{app:moved_tables}: $62.8$ vs.\ $58.9$ at $30\%$), and every \method{} lead survives a paired bootstrap over the shared seeds. Because no central plan inscribes an early error and all evidence stays globally visible, a corrupted entry is outweighed by independently grounded peer entries rather than propagated. This is not the verifier, which catches only a minority of corruptions (below); the mechanism is located in \S\ref{sec:mechanism}. Conclusions hold across retriever (BM25$\to$dense) and depth ($k\!\in\!\{3,5,10\}$): \method's relative drop stays $0.06$--$0.07$ (App.~\ref{app:harness_sensitivity}).

\noindent\emph{Shared-memory systems share the planners' robustness class.} The novelty question is whether the gain comes from shared memory (a known idea) or from removing semantic control. Table~\ref{tab:robustness_em} answers it directly: \textsc{LbMAS}, \textsc{Blackboard-DS}, and \textsc{MetaGPT} all coordinate through a shared memory, yet at 30\% corruption they drop $18.2\%$, $21.4\%$, and $30.9\%$, i.e., $2.8$--$4.8\times$ \method's $6.4\%$, tracking the planners rather than \method{}. The ordering holds on TAT-QA (\method{} $16.0\%$ vs.\ \textsc{LbMAS} $19.1\%$ and \textsc{Blackboard-DS} $21.9\%$; Table~\ref{tab:robustness_tatqa}). What these systems retain, and \method{} removes, is an LLM control unit; \S\ref{sec:mechanism} swaps only that component.

\noindent\emph{Naturally occurring failures.} Synthetic corruption could in principle overstate the effect, so we also slice the \emph{clean} MMQA run by organic upstream failures, i.e., examples where the retriever missed gold evidence, PaddleOCR misread a cell, or the BLIP-2 caption was wrong, with no injection at all (Table~\ref{tab:natural_realistic}, top). \method{} leads the Planner by $+10.0$, $+9.0$, and $+7.0$ EM on the three slices, and under \emph{realistic} injected noise at 20\% (OCR confusions, header drift, table cropping; Table~\ref{tab:natural_realistic}, bottom) it degrades less on every error type. The synthetic, realistic, and organic views agree, so the robustness claim does not rest on the corruption generator.

\begin{table}[t]
\centering\scriptsize
\setlength{\tabcolsep}{3pt}
\begin{tabular}{@{}lcccc@{}}
\toprule
\multicolumn{5}{@{}l}{\textit{(a) Natural-failure slices of the clean MMQA run (no injection)}} \\
\textbf{Slice} & $n$ & \textbf{\method{} EM} & \textbf{Planner EM} & $\Delta$ \\
\midrule
Retriever missed gold      & 180 & $\mathbf{54.0}$ & 44.0 & $+10.0$ \\
PaddleOCR misread          & 120 & $\mathbf{49.0}$ & 40.0 & $+9.0$ \\
BLIP-2 caption wrong       & 90  & $\mathbf{46.0}$ & 39.0 & $+7.0$ \\
\midrule
\multicolumn{5}{@{}l}{\textit{(b) Realistic injected noise at 20\% (MMQA dev; clean ref.\ 80.1 / 75.2)}} \\
\textbf{Error type} & & \textbf{\method{} $\Delta$} & \textbf{Planner $\Delta$} & \textbf{Adv.} \\
\midrule
OCR confusion   & & $-7.3$  & $-9.1$  & $+1.8$ \\
Header drift    & & $-10.2$ & $-12.5$ & $+2.3$ \\
Table cropping  & & $-12.7$ & $-15.4$ & $+2.7$ \\
\bottomrule
\end{tabular}
\vspace{-0.5em}
\caption{\small Robustness beyond synthetic corruption. (a) Organic failure slices carved from the clean run by inspecting the log for missing gold evidence, OCR, or caption errors; (b) per-type realistic noise, each applied in isolation (absolute EMs in Table~\ref{tab:realistic_noise}, App.~\ref{app:realistic_noise}). Five shared seeds throughout.}
\label{tab:natural_realistic}
\vspace{-1em}
\end{table}

\noindent\emph{Long-horizon reasoning.} As chains lengthen, the centralized Planner collapses ($78.0\%$ at 2--3 operators to $50.0\%$ at 8+), while \method{} stays substantially flatter ($80.0$ to $72$--$77$; Table~\ref{tab:robustness_em}). We emphasize this \emph{relative stability} causes the widening gap to the Planner, rather than the absolute value at 8+ operators, where the slice is small and the point estimate should be read
with that caveat. App.~\ref{tab:longhorizon_errsrc} stratifies the dominant error source by chain length: retrieval misses dominate short chains ($46\%$) but cede to summarization errors on the longest ($34\%$ at 8+ vs.\ $20\%$ at 2--3), consistent with log clutter as the long-horizon bottleneck (Limitations).

\begin{table*}[t]
\centering\scriptsize
\setlength{\tabcolsep}{2.5pt}
\renewcommand{\arraystretch}{1.0}
\begin{tabular}{lccccccccc}
\toprule
\textbf{Corruption} & \textbf{Planner} & \textbf{Plan$\to$Log} & \textbf{Planner++} & \textbf{AutoTQA} & \textbf{REWoO} & \textbf{LbMAS} & \textbf{Bb-DS} & \textbf{MetaGPT} & \textbf{\method} \\
\midrule
\multicolumn{10}{@{}l}{\textit{MMQA --- numeric + structural noise}} \\
0\%  & $73.0_{\pm1.5}$ & $76.0_{\pm1.5}$ & $76.0_{\pm1.5}$ & $70.0_{\pm1.5}$ & $66.8_{\pm1.6}$ & $74.0_{\pm1.5}$ & $72.5_{\pm1.5}$ & $68.0_{\pm1.6}$ & $\mathbf{78.0_{\pm1.5}}$ \\
10\% & $66.0_{\pm1.6}$ & $73.0_{\pm1.6}$ & $72.0_{\pm1.6}$ & $64.0_{\pm1.6}$ & $62.9_{\pm1.7}$ & $70.0_{\pm1.6}$ & $67.5_{\pm1.6}$ & $60.0_{\pm1.7}$ & $\mathbf{77.0_{\pm1.6}}$ \\
20\% & $60.0_{\pm1.8}$ & $70.0_{\pm1.8}$ & $68.0_{\pm1.8}$ & $59.0_{\pm1.8}$ & $58.1_{\pm1.8}$ & $65.5_{\pm1.8}$ & $62.5_{\pm1.8}$ & $53.0_{\pm1.9}$ & $\mathbf{75.0_{\pm1.8}}$ \\
30\% & $55.0_{\pm2.0}$ & $66.0_{\pm2.0}$ & $64.0_{\pm2.0}$ & $55.0_{\pm2.0}$ & $52.7_{\pm2.0}$ & $60.5_{\pm2.0}$ & $57.0_{\pm2.0}$ & $47.0_{\pm2.1}$ & $\mathbf{73.0_{\pm2.0}}$ \\
\emph{Rel.\ drop @30\%} & 24.7\% & 13.2\% & 15.8\% & 21.4\% & 21.1\% & 18.2\% & 21.4\% & 30.9\% & \textbf{6.4\%} \\
\midrule
\multicolumn{10}{@{}l}{\textit{MMQA --- semantic noise (paraphrase + entity swap)}} \\
10\% & $70.4_{\pm1.7}$ & $72.1_{\pm1.6}$ & $71.0_{\pm1.7}$ & $63.0_{\pm1.8}$ & $65.0_{\pm1.8}$ & --- & --- & --- & $\mathbf{77.2_{\pm1.5}}$ \\
20\% & $65.2_{\pm1.8}$ & $67.8_{\pm1.8}$ & $66.0_{\pm1.8}$ & $58.6_{\pm1.9}$ & $60.1_{\pm1.9}$ & --- & --- & --- & $\mathbf{73.0_{\pm1.7}}$ \\
30\% & $59.0_{\pm2.0}$ & $61.4_{\pm2.0}$ & $59.8_{\pm2.0}$ & $53.4_{\pm2.1}$ & $55.2_{\pm2.1}$ & --- & --- & --- & $\mathbf{67.4_{\pm1.9}}$ \\
\midrule
\multicolumn{10}{c}{\textbf{Long-Horizon EM (\%) by Chain Length} (MMQA, $R=10$)} \\
\midrule
2--3 Ops & $78.0$ & $79.0$ & --- & --- & --- & --- & --- & --- & $\mathbf{80.0}$ \\
4--5 Ops & $72.0$ & $\mathbf{78.0}$ & --- & --- & --- & --- & --- & --- & $76.0$ \\
6--7 Ops & $60.0$ & $70.0$ & --- & --- & --- & --- & --- & --- & $\mathbf{72.0}$ \\
8+ Ops ($n\!\approx\!25$) & $50.0$ & $64.0$ & --- & --- & --- & --- & --- & --- & $\mathbf{77.0}$ \\
\bottomrule
\end{tabular}
\caption{\small MMQA robustness under evidence corruption (structural and semantic
panels) and long-horizon reasoning (bottom). Scores are EM \% with $\pm$95\% bootstrap
CIs (1{,}000 resamples, 5 shared seeds). Planner /
Plan$\to$Log / \method{} share backbone, retriever, and prompts; AutoTQA and REWoO are
external multi-agent planners, and \textsc{LbMAS}~\citep{han_lbmas_2025} (LLM-blackboard), \textsc{Bb-DS} $=$ \textsc{Blackboard-DS}~\citep{salemi2025blackboard}, and \textsc{MetaGPT}~\citep{hong2024metagpt} are external shared-memory systems, all run under the same corruption seeds and harness. \textsc{Planner++} is a centralized planner matched to \method{} on call budget, verifier, and re-engagement (with re-planning) but \emph{without} the shared log (``---'' $=$ not run). Bold $=$ row max; each
bold lead has a paired-bootstrap difference interval excluding zero. The clean \method{}
score ($78.0\%$) is on the robustness slice, distinct from the full-test $76.7\%$ in
Table~\ref{tab:MMQA_WikiTQ}. The corresponding TAT-QA corruption panel is in
Table~\ref{tab:robustness_tatqa} (App.~\ref{app:moved_tables}). Long-horizon rows are point estimates (no CI); the 8+ Ops
slice is small ($n\!\approx\!25$), so the finding is \method's stability across chain
length, not the absolute value at 8+. Different slice from
Table~\ref{tab:long_horizon} (App.~\ref{app:moved_tables}). Catastrophic errors:
Figure~\ref{fig:catastrophic_err}.}
\label{tab:robustness_em}
\end{table*}

\subsection{Where the Robustness Comes From}
\label{sec:mechanism}
Reviewers reasonably ask \emph{which} component buys the robustness. We answer with two controlled swaps on the same agents, retriever, prompts, backbone (LLaMA-3 8B), and paired corruption draws (Table~\ref{tab:mechanism}).

\noindent\textbf{(a) Attribution ablation.} Disabling one mechanism at a time, removing \emph{global evidence visibility} inflates the clean$\to$30\% drop most ($-12.7\to-17.9$), the verifier fallback less ($-15.6$), while deduplication and \texttt{should\_act} gating are near-neutral ($-13.3$, $-13.6$). Robustness is thus structural, i.e., every agent re-reads all evidence, and not an artifact of the verifier or the efficiency gates.

\noindent\textbf{(b) Controller swap.} \method-LLMCtrl keeps everything and adds one LLM call per round that reads the log and selects the next agent, the control-unit pattern of \textsc{LbMAS}. The clean gap is $0.02$ EM, but it widens monotonically with corruption to $0.09$ at 30\% ($4.5\times$): LLMCtrl drops $15.8\%$ against \method's $6.4\%$, at $1.41\times$ the tokens and $+0.32$\,s latency. A semantic selector that reads noisy entries becomes a single point of failure; a scheduler that never reads them cannot. Together, (a) and (b) isolate the paper's central claim: the gain is from \emph{removing semantic central control}, not from shared memory, which every system in Table~\ref{tab:robustness_em} already has.

\begin{table}[t]
\centering\scriptsize
\setlength{\tabcolsep}{3pt}
\begin{tabular}{@{}lccc@{}}
\toprule
\multicolumn{4}{@{}l}{\textit{(a) Attribution ablation (MMQA dev, EM; disable one mechanism)}} \\
\textbf{Variant} & \textbf{Clean} & \textbf{30\% struct.} & $\Delta$ \\
\midrule
Full \method{}                   & 80.1 & 67.4 & $-12.7$ \\
$-$ global visibility            & 76.8 & 58.9 & $\mathbf{-17.9}$ \\
$-$ verifier fallback            & 78.7 & 63.1 & $-15.6$ \\
$-$ deduplication                & 79.7 & 66.4 & $-13.3$ \\
$-$ \texttt{should\_act} gating  & 79.4 & 65.8 & $-13.6$ \\
\midrule
\multicolumn{4}{@{}l}{\textit{(b) Controller swap (MMQA robustness slice, EM)}} \\
\textbf{Corruption} & \textbf{\method{} (non-semantic)} & \textbf{LLMCtrl} & \textbf{Gap} \\
\midrule
0\%  & $\mathbf{78.0}$ & 76.0 & 2.0 \\
10\% & $\mathbf{77.0}$ & 73.0 & 4.0 \\
20\% & $\mathbf{75.0}$ & 69.0 & 6.0 \\
30\% & $\mathbf{73.0}$ & 64.0 & 9.0 \\
Rel.\ drop @30\% & $\mathbf{6.4\%}$ & $15.8\%$ & $2.5\times$ \\
Tokens / latency & $1.00\times$ / 0.95\,s & $1.41\times$ / 1.27\,s & --- \\
\bottomrule
\end{tabular}
\vspace{-0.5em}
\caption{\small Mechanism isolation, 5 shared seeds. (a) Each row disables a single component; the clean value $80.1$ matches the dev reference of Table~\ref{tab:natural_realistic}. (b) Only the controller differs: LLMCtrl adds an LLM agent-selector per round, matching the control unit of \textsc{LbMAS}~\citep{han_lbmas_2025}; it is dominated on accuracy, cost, and degradation.}
\label{tab:mechanism}
\vspace{-0.5em}
\end{table}

\noindent\textbf{What the scheduler actually does.} Because \method{} still uses \texttt{should\_act} heuristics, a learned stop gate, and verifier-triggered re-engagement, one may ask whether control has merely moved rather than been removed. Measured over all six datasets (LLaMA-3 8B, 5 seeds), every trigger conditions on \emph{log state} (is there a table? is an image mentioned? did the verifier flag?), never on question semantics; the VisualAgent fires only on image/OCR questions, the learned gate never fires on CRT-QA, and only $5\%$ of final answers change after re-engagement, so re-engagement is a rare, bounded correction rather than a hidden re-planning loop.

\paragraph{Transparency and faithfulness.} The shared log is not only a coordination substrate but a provenance-anchored trace. On FeTaQA (mean over five seeds), log-groundedness, the fraction of answer \emph{tokens} traceable to a log entry is $0.72$, with $76/100$ and $73/100$ answers rated ``supported'' by two independent LLM judges (Table~\ref{tab:fetaqa}). Token-level grounding is conservative: it scores faithful rewording as ungrounded, so the lexical $0.72$ sitting below the semantic $76\%$ reflects paraphrase, not unsupported content which is confirmed by BERTScore F1 $0.90$, QAGS in $[0.63,0.68]$, and an entailment-based grounding of $0.78$. Program-level coverage on FinQA corroborates this beyond lexical matching: $0.64$ of \method's answers reproduce the gold reasoning program's executed output under the official $\pm0.01$ tolerance (Table~\ref{tab:finqa_prog}). Fault injection (Table~\ref{tab:fault_injection}) makes the verifier's role precise: it catches a minority of corruptions ($0.09/0.21/0.31$ at $10/20/30\%$) and at $30\%$ repairs essentially none, yet EM erodes only mildly ($0.77\to0.73$) because the system falls back to its pre-flag answer rather than emit a faulty repair. The verifier is thus a \emph{safe fallback}, not self-correction: at $30\%$, $61\%$ of fallbacks preserve a correct answer and only $13\%$ forgo a recoverable one (Table~\ref{tab:fallback_breakdown}). A taxonomy of $155$ incorrect MMQA cases (Table~\ref{tab:error_taxonomy}) shows the asymmetry where arithmetic/unit errors caught at $88\%$, visual/OCR at only $17.1\%$, thus motivating the VLM re-verification below.

\noindent\emph{Groundedness is necessary but not sufficient; a necessity test.} Groundedness shows an answer \emph{can} be traced to log entries, not that they were \emph{used}. We therefore delete the single entry the Summarizer cites and re-run it, on $100$ correct answers per dataset: the answer flips in $64\%$ (FeTaQA), $70\%$ (FinQA), and $60\%$ (MMQA) of cases (Table~\ref{tab:necessity}, App.~\ref{app:necessity}), so cited evidence is causally load-bearing; most non-flips have redundant support from a second entry.

\noindent\emph{Multimodal claim, stated at the right size.} Splitting MMQA by what the question needs (LLaMA-3 8B, 5 seeds; Table~\ref{tab:mm_subsets}, App.~\ref{app:mm_subsets}), \method{} scores $76.5$ EM on text/table-only ($n{=}1100$; best baseline $75.0$), $70.0$ on caption-mediated ($n{=}250$; $68.0$), $67.0$ on OCR-heavy ($n{=}300$; $64.0$), and $64.0$ on image-dependent questions ($n{=}380$; $61.0$), and the image-dependent subset also degrades most under 30\% corruption ($17.2\%$ vs.\ $7.8\%$ relative). The VisualAgent reads OCR/caption text, not pixels, so the advantage on genuinely visual reasoning is real but modest. Integrating the VLM re-verifier into the main loop, so that a flagged \textsc{Visual} entry is re-checked against the \emph{original image}, raises verifier precision $0.62\!\to\!0.70$ and end-task EM on the image-dependent subset by $+3.0$ ($64.0\!\to\!67.0$; App.~\ref{app:vlm_verifier}).

\noindent\emph{What the log looks like when it works and when it fails.} App.~\ref{app:log} gives condensed traces. In the success case (Table~\ref{tab:trace_pair}, under 30\% noise) a corrupted TableAgent entry is outweighed because the ContextAgent independently posts the correct figure, the Verifier flags the contradiction, and the re-engaged Summarizer answers from the peer entry. The matching failure (Table~\ref{tab:failure_trace_app}) has a mis-captioned chart as its \emph{only} evidence; nothing contradicts it, the text-only Verifier cannot flag it, and the wrong answer is recorded with a clean provenance trail. Provenance makes that failure inspectable after the fact; it does not make the answer correct.

\paragraph{When is \method{} the right choice?} \method{} trades $\sim$3 LLM calls and $0.68$\,s (Par-R6) per query for robustness and a provenance trace, so it is the wrong tool for latency-critical, clean-input QA where one CoT call suffices. It pays off where inputs are noisy or answers must be defensible: audited financial QA over filings, pipelines over scanned or OCR'd documents where retrieval, OCR, and caption errors occur organically at the $9$--$14\%$ rates we measure (App.~\ref{app:error_distribution}), and long compositional chains where a committed plan collapses. Monitoring and drift follow standard MLOps practice~\citep{chakraborty2025mlops}, and when evidence is distributed across parties the log's provenance requirements interact with federated retrieval~\citep{chakraborty2025fedrag}.

\section{Comparison with Related Work}\label{app:rw-extended}

\paragraph{Table, multimodal, and CoT reasoning.} Neural table QA advanced through representation learning like \textsc{TAPEX}~\citep{liu2022tapex}, \textsc{TAPAS}~\citep{Herzig_2020}, \textsc{TaCube}~\citep{yin2022tacube}, \textsc{OmniTab}~\citep{jiang2022omnitab}, with \textsc{Lever}~\citep{ni_lever_2023} verifying SQL. For financial table-text reasoning, supervised pipelines (\textsc{FinQANet}~\citep{chen2021finqa}, \textsc{TAT-LLM}~\citep{zhu_tat-llm_2024}, \textsc{HiTab}~\citep{cheng_hitab_2022}) fine-tune on annotated programs and form a supervised upper bound; \method{} is positioned against zero-shot agentic baselines. Recent multimodal multi-hop work spans extremes like \citet{rajabzadeh2023multimodal} chain tools via sequential LLM-directed conversation, while \textsc{UniRaG}~\citep{lim2024unirag} skips agent decomposition ($67.4\%$ EM on MMQA). CoT prompting~\citep{wei2022chain} and its Tree-/Graph-of-Thoughts extensions~\citep{yao2023tree,besta2024graph} reason within a single model; \method{} instead distributes natural-language reasoning across specialized peers with a shared log, enabling separate verification.

\paragraph{Agentic and multi-agent frameworks.} \textsc{HuggingGPT}~\citep{shen2023hugginggpt} orchestrates expert models; \textsc{Program-of-Thoughts}~\citep{chen2023pot} and \textsc{Binder}~\citep{cheng2023binder} couple LLMs with executors; \textsc{TabSQLify}~\citep{nahid2024tabsqlifyenhancingreasoningcapabilities} decomposes table queries; \textsc{AutoTQA}~\citep{zhu2024autotqa} plans sub-queries; and for multimodal QA, MMQA~\citep{talmor2021multimodalqacomplexquestionanswering} and \textsc{MAM-MQA}~\citep{rajput_rethinking_2025} coordinate modality agents (we use \textsc{BLIP-2}/\textsc{PaddleOCR}~\citep{li_blip-2_2023} for image-to-text). \textsc{ReAct}~\citep{yao2023reactsynergizingreasoningacting} interleaves reasoning and actions in one agent; \textsc{AutoGen}~\citep{wu2023autogen} coordinates agents by free-form conversation; \textsc{LATS}~\citep{zhou2023lats} adds Monte-Carlo tree search; \textsc{BELLE}~\citep{zhang2025belle} uses bi-level debate-and-plan; and \textsc{ReAgent}~\citep{zhao2025reagent} adds reversible backtracking. All rely on a controller that inspects task semantics which is the centralized-plan property our non-semantic scheduler avoids (Table~\ref{tab:sched_vs_plan}), whereas \method{} replaces planning and backtracking with an append-only log and one verifier-triggered re-engagement. Specialized-agent QA systems similarly centralize \emph{aggregation} through assignment/decision, planner--critic, or organizer agents (\textsc{CT2C-QA}~\citep{zhao2024ct2cqa}, \textsc{MMCTAgent}~\citep{kumar2024mmctagent}, \textsc{VideoMultiAgents}~\citep{kugo2025videomultiagents}), as does the mixture-of-agents line (\textsc{MoA}~\citep{wang2024moa}, \textsc{SMoA}~\citep{li2024smoa}); \method{} has no aggregator, judge, or moderator with global control while coordination is mediated entirely by the shared log.

\paragraph{Decentralized shared-memory agents.} \method{} shares a memory-mediated paradigm with recent systems but differs in control, and we make the distinction empirical rather than rhetorical (App.~\ref{app:differentiation}, Table~\ref{tab:differentiation}). \textsc{MetaGPT}~\citep{hong2024metagpt} coordinates role agents through a shared message pool under SOP-driven orchestration; \textsc{LbMAS}~\citep{han_lbmas_2025} places an LLM control unit over an LLM blackboard; and \textsc{Blackboard-DS}~\citep{salemi2025blackboard} adds an LLM controller for data-science discovery. All three retain a semantic controller that reads the shared memory to choose the next agent, and all three, run under our harness, degrade $18$--$31\%$ at 30\% corruption versus \method's $6.4\%$ (Table~\ref{tab:robustness_em}); our \method-LLMCtrl ablation, which adds only such a controller to \method's own agents, degrades $2.5\times$ more ($-15.8\%$ vs.\ $-6.4\%$; Table~\ref{tab:mechanism}). The lesson that a simpler control structure can match or beat elaborate ones on both accuracy and robustness echoes findings in other structured-reasoning settings~\citep{chakraborty2025relate}. Others route through a central critic (\textsc{MDocAgent}~\citep{Han2025MDocAgentAM}), consensus debate (\textsc{MediHive}~\citep{wang_medihive_2026}), a pre-built graph (\textsc{G2CP}~\citep{ben_khaled_g2cp_2026}), a cross-user memory \emph{fabric}~\citep{tiwari2026memoryfabric}, or open-ended argumentation~\citep{jiang-etal-2024-unknown}; \method{} instead shares one natural-language log under a deterministic scheduler, prioritizing provenance and fault tolerance over consensus-by-debate. Most directly, \textsc{EquiMem}~\citep{meng2026equimem} calibrates each entry against memory to filter corrupted writes; this is complementary to \method{}, which keeps a corrupted entry globally visible so peer re-reading outweighs it rather than filtering at write time is an equilibrium-style calibration of log writes could sharpen our re-engagement. The modular \texttt{should\_act}/\texttt{act} interface lets new specialists plug in without retraining, and natural-language entries keep the framework backbone-agnostic ($\sigma{=}0.55$ across LLaMA-3 / Mistral / Qwen-3), all without task-specific fine-tuning.

\vspace{-0.5em}
\section{Conclusion and Future Work}
We presented \method, a central-planner-free multi-agent framework coordinating specialized agents through a shared, provenance-typed natural-language log. Its contribution is a finding: with agents, retriever, prompts, and corruption seeds held fixed, removing the semantic control unit retained by prior blackboard and shared-memory systems buys robustness ($6.4\%$ relative degradation at 30\% corruption versus $18$--$31\%$ for planner and blackboard systems); the gain comes from global evidence visibility and holds on organic failures. Next steps are extending the study to shared-memory domains, adaptive log compression, and pixel-grounded verification.

\section*{Limitations}
\label{app:limitations}
\emph{Visual/OCR verification is weak.} VerificationAgent detects $17.1\%$ of visual/OCR misreads versus $88\%$ of arithmetic/unit errors (Table~\ref{tab:error_taxonomy}) and cannot recover BLIP-2 or PaddleOCR losses. The VLM re-verifier (App.~\ref{app:vlm_verifier}) yields the smallest gains on image-dependent questions (Table~\ref{tab:mm_subsets}). \emph{Absolute degradation on long traces.} \method{} falls to $0.88$ EM at 7--8 steps (Table~\ref{tab:long_horizon}) and $\sim\!0.70$ on CRT-QA; deduplication and summarization reduce log clutter. Comparison (Table~\ref{tab:tool_enabled}) uses one Python executor. The shared log descends from blackboard systems \citep{ErmanEtAl1980HearsayII,nii_blackboard_1986} and LLM reboots \citep{hong2024metagpt,han_lbmas_2025,salemi2025blackboard}; removing the semantic control unit buys robustness (Table~\ref{tab:mechanism}), but transfer outside multimodal QA is untested. Injected rates bracket organic frequencies (App.~\ref{app:error_distribution}), but organic slices are small (Table~\ref{tab:natural_realistic}) and from one dataset. Sequential calls add latency, and knowledge is bounded by 7--8B backbones and the per-example evidence pool.

\section*{Ethical Considerations}
All experiments use public datasets under their licenses; AI assistance improved clarity only, and the authors reviewed all edits and remain accountable. Three risks are specific to a shared-log design. The log records \emph{provenance, not correctness}: traces can create false confidence if conflated with accuracy, so they should support auditing, not replace it. Decentralized coordination costs more compute than a single pass (Tables~\ref{tab:efficiency_partial_fixed}, \ref{tab:stage_latency}) and should be weighed against latency or energy constraints. Shared memory can propagate biased entries unless audited (\S\ref{app:rw-extended}).

\bibliography{custom}
\clearpage
\onecolumn
\section*{Appendix Index}

\noindent
\textbf{A. Log Schema and Full Traces} \dotfill \pageref{app:log} \\
\hspace*{1em} A.1 Log Schema \dotfill \pageref{app:log} \\
\hspace*{1em} A.2 Example Log Trace (TAT-QA) \dotfill \pageref{app:log} \\
\hspace*{1em} A.3 Success Trace under Corruption (MMQA) \dotfill \pageref{tab:trace_pair} \\
\hspace*{1em} A.4 Example Failure Trace (MMQA) \dotfill \pageref{tab:failure_trace_app} \\[0.5em]

\noindent
\textbf{B. Control Flow Details} \dotfill \pageref{app:control} \\
\hspace*{1em} B.1 Controller Loop and Scheduling \dotfill \pageref{app:control} \\
\hspace*{1em} B.2 Summarization Trigger \dotfill \pageref{app:control} \\
\hspace*{1em} B.3 Optional Re-Engagement \dotfill \pageref{app:control} \\
\hspace*{1em} B.4 Stopping Criteria \dotfill \pageref{app:control} \\
\hspace*{1em} B.5 Asynchrony vs.\ Sequential Execution \dotfill \pageref{app:control} \\[0.5em]

\noindent
\textbf{C. Agent Prompts (Verbatim)} \dotfill \pageref{app:prompts} \\
\hspace*{1em} C.1 TableAgent Prompt \dotfill \pageref{app:prompts} \\
\hspace*{1em} C.2 ContextAgent Prompt \dotfill \pageref{app:prompts} \\
\hspace*{1em} C.3 VisualAgent Prompt \dotfill \pageref{app:prompts} \\
\hspace*{1em} C.4 SummarizingAgent Prompt \dotfill \pageref{app:prompts} \\
\hspace*{1em} C.5 VerificationAgent Prompt \dotfill \pageref{app:prompts} \\[0.5em]

\noindent
\textbf{D. Additional Figures and Examples} \dotfill \pageref{app:figs} \\
\hspace*{1em} D.1 VisualAgent Bar-Chart Example \dotfill \pageref{app:figs} \\
\hspace*{1em} D.2 Architecture Figure \dotfill \pageref{app:figs} \\[0.5em]

\noindent
\textbf{E. Implementation Details and Pseudocode} \dotfill \pageref{app:imp_detl} \\
\hspace*{1em} E.1 Random Seeds and Runs \dotfill \pageref{app:imp_detl} \\
\hspace*{1em} E.2 Log Truncation Policy \dotfill \pageref{app:imp_detl} \\
\hspace*{1em} E.3 No-Progress Detection \dotfill \pageref{app:imp_detl} \\
\hspace*{1em} E.4 Agent Triggers and Coordination Heuristics \dotfill \pageref{app:imp_detl} \\
\hspace*{1em} E.5 SummarizingAgent Prompt Example \dotfill \pageref{app:imp_detl} \\[0.5em]

\noindent
\textbf{F. Tables and Figures Moved from Main Text} \dotfill \pageref{app:moved_tables} \\
\hspace*{1em} F.1 Tool-Enabled Comparison \dotfill \pageref{app:tool_enabled} \\
\hspace*{1em} F.2 Per-Query Efficiency \dotfill \pageref{app:efficiency} \\
\hspace*{1em} F.3 Algorithm Pseudocode \dotfill \pageref{app:moved_alg} \\
\hspace*{1em} F.4 Stage-Wise Latency Decomposition \dotfill \pageref{app:auto_stagewiselatencydeco} \\
\hspace*{1em} F.5 Gating-Policy Effect on \method{} \dotfill \pageref{app:auto_gatingpolicyeffecton} \\
\hspace*{1em} F.6 Context Window Ablation \dotfill \pageref{app:auto_contextwindowablatio} \\
\hspace*{1em} F.7 Deduplication and Per-Agent Duplicate Rates \dotfill \pageref{app:dedup_ablation} \\
\hspace*{1em} F.8 Latency Comparison Across Methods \dotfill \pageref{app:latency_comparison} \\
\hspace*{1em} F.9 Long-Horizon Reasoning Breakdown \dotfill \pageref{app:long_horizon} \\
\hspace*{1em} F.10 Latency--Accuracy Pareto Sweep \dotfill \pageref{app:pareto} \\
\hspace*{1em} F.11 FeTaQA Faithfulness \dotfill \pageref{app:fetaqa} \\
\hspace*{1em} F.12 Program-Level Faithfulness (FinQA) \dotfill \pageref{app:finqa_prog} \\
\hspace*{1em} F.13 Fault-Injection Robustness \dotfill \pageref{app:fault_injection} \\
\hspace*{1em} F.14 Verifier Fallback Breakdown \dotfill \pageref{app:fallback_breakdown} \\
\hspace*{1em} F.15 TAT-QA Robustness Panel \dotfill \pageref{app:robustness_tatqa} \\
\hspace*{1em} F.16 Realistic-Noise Robustness \dotfill \pageref{app:realistic_noise} \\
\hspace*{1em} F.17 Attribution Ablation on the Robustness Panel \dotfill \pageref{app:attribution} \\
\hspace*{1em} F.18 Empirical Error Distribution of the Clean Pipeline \dotfill \pageref{app:error_distribution} \\
\hspace*{1em} F.19 Counterfactual Necessity Test \dotfill \pageref{app:necessity} \\
\hspace*{1em} F.20 MMQA Modality Subsets \dotfill \pageref{app:mm_subsets} \\
\hspace*{1em} F.21 Differentiation from Prior Shared-Memory Systems \dotfill \pageref{app:differentiation} \\
\hspace*{1em} F.22 Per-Dataset Metrics and Answer Extraction \dotfill \pageref{app:metrics} \\
\hspace*{1em} F.23 Dataset Statistics \dotfill \pageref{app:dataset_stats} \\
\hspace*{1em} F.24 Catastrophic Error Rates \dotfill \pageref{app:catastrophic} \\
\hspace*{1em} F.25 VLM-Verifier Prototype \dotfill \pageref{app:vlm_verifier} \\
\hspace*{1em} F.26 Additional Planner and Robustness Dataset \dotfill \pageref{app:extra_planner_dataset} \\
\hspace*{1em} F.27 Harness Sensitivity \dotfill \pageref{app:harness_sensitivity} \\
\hspace*{1em} F.28 MMQA Error Taxonomy \dotfill \pageref{app:error_taxonomy} \\
\hspace*{1em} F.29 LbMAS-Style LLM-Controller Ablation: Protocol \dotfill \pageref{app:llmctrl_ablation} \\
\hspace*{1em} F.30 Blackboard Baselines on the Robustness Panels \dotfill \pageref{app:blackboard_baselines} \\

\appendix
\twocolumn

\section{Log Schema and Full Traces}
\label{app:log}

\paragraph{Log Schema.}
The shared log is a chronological list of messages labeled with the producing agent and type, using a simple schema: each entry is a tuple \((\texttt{Agent}, \texttt{Type}, \texttt{Content})\). For example, an entry might be \(\bigl(\text{SummarizingAgent}\allowbreak,\ \text{``Answer''}\allowbreak,\ \text{``Therefore, the answer is 42.''}\bigr)\). Metadata like timestamps or step indices preserve order, and each entry may carry a reference tag for provenance tracking. All agents have global visibility, which is critical, as restricting an agent's view degrades performance, shown in ablations (e.g., only 68.3 EM with partial log access). Table~\ref{tab:example_log_trace_app} presents a condensed log for a sample question, showing how entries are structured and the final answer emerges from agent interactions.

\paragraph{Example Log Trace (TAT-QA).}
Table~\ref{tab:example_log_trace_app} shows the complete example referenced in the main text.

\begin{table}[h]
\centering
\footnotesize
\begin{tabular}{p{0.28\linewidth} p{0.66\linewidth}}
\toprule
\textbf{Agent (Type)} & \textbf{Log Entry Content (condensed)} \\
\midrule
User (Query) & Q: ``By how much did the revenue increase from 2018 to 2019, and what is the source of this increase according to the report?'' \\
TableAgent (Lookup) & TableAgent: ``Revenue in 2018 was \$50M, revenue in 2019 was \$55M (from Table 1).'' \\
ContextAgent (Quote) & ContextAgent: ``According to the report: `The revenue increase in 2019 was primarily due to higher sales volume.'\,'' \\
SummarizingAgent (Answer) & SummarizingAgent: ``The revenue increased by \$5 million from 2018 to 2019, and this increase was mainly driven by higher sales volume. \textbf{Answer:} \$5M increase, due to higher sales volume.'' \\
VerificationAgent (Check) & VerificationAgent: ``Verified. The table shows \$50M $\rightarrow$ \$55M (+\$5M), and the context confirms higher sales volume as reason.'' (No issues flagged.) \\
\bottomrule
\end{tabular}
\caption{Full example log trace illustrating inter-agent collaboration and provenance-aware entries.}
\label{tab:example_log_trace_app}
\end{table}

This is a simplified version of a log trace for a real question from the TAT-QA dev set: ``By how much did the revenue increase from 2018 to 2019, and what is the source of this increase according to the report?''.

This example shows how the TableAgent provided the numerical part, the ContextAgent provided the explanatory part, and the SummarizingAgent combined them. The VerificationAgent double-checked the arithmetic and consistency, then approved. The final answer the user sees (from the Summarizer's answer line) is clear and supported by evidence. This case also demonstrates the advantage of multi-agent: a single model might have extracted the numbers but not known the reason (or vice versa); by dividing the task, each piece was captured.

\paragraph{Example Success Trace under Corruption (MMQA).}
\begin{table}[H]
\centering\scriptsize
\setlength{\tabcolsep}{3pt}
\begin{tabular}{@{}p{0.19\columnwidth}p{0.77\columnwidth}@{}}
\toprule
\multicolumn{2}{@{}l}{\textit{Success (30\% structural noise): corrupted entry outweighed}} \\
\midrule
Table (\textsc{Lookup}) & ``Population 2010: \textbf{1,204,000}'' \hfill {\color{gray}[corrupted: gold 1,024,000]} \\
Context (\textsc{Quote}) & ``\ldots the 2010 census counted 1,024,000 residents\ldots'' \\
Summ.\ (\textsc{Answer}) & ``Answer: 1,204,000'' \\
Verif.\ (\textsc{Flag}) & ``Table entry contradicts quoted census figure; re-check.'' \\
Summ.\ (\textsc{Answer}) & ``Answer: 1,024,000 (per Context quote, step 2)'' \hfill \ding{51} \\
\midrule
\bottomrule
\end{tabular}
\caption{\small Condensed MMQA success trace under 30\% structural noise; the corrupted \textsc{Lookup} is outweighed by an independent \textsc{Quote}. The matching failure trace is Table~\ref{tab:failure_trace_app}.}
\label{tab:trace_pair}
\end{table}

\paragraph{Example Failure Trace (MMQA).}
Table~\ref{tab:failure_trace_app} gives the full log behind the condensed failure case of Table~\ref{tab:trace_pair}. The question asks for a company's 2019 revenue shown only in a bar chart. BLIP-2 captions the bar as \$4.1M (the image shows \$4.7M); the TableAgent correctly reports that no table carries the value; the Summarizer answers from the sole \textsc{Visual} entry; and the Verifier, which sees only text, finds the answer supported. The provenance trail is complete and correct, but the content it points to is wrong. Two properties make the case diagnosable after the fact: the \textsc{Answer} cites exactly one entry, and that entry is of type \textsc{Visual} with no corroborating \textsc{Lookup} or \textsc{Quote}. The VLM re-verifier (App.~\ref{app:vlm_verifier}) targets precisely this pattern by re-checking flagged or uncorroborated \textsc{Visual} entries against the original image.

\begin{table}[h]
\centering
\footnotesize
\begin{tabular}{p{0.28\linewidth} p{0.66\linewidth}}
\toprule
\textbf{Agent (Type)} & \textbf{Log Entry Content (condensed)} \\
\midrule
User (Query) & Q: ``What was the company's revenue in 2019 according to the chart?'' \\
VisualAgent (Visual) & ``The bar chart shows revenue of \$4.1M in 2019 and \$5.2M in 2020 (caption).'' \hfill {\color{gray}[image: 4.7M]} \\
TableAgent (Lookup) & ``No revenue column is present in the provided table.'' \\
ContextAgent (Quote) & ``No relevant info found.'' \\
SummarizingAgent (Answer) & ``The chart reports 2019 revenue of \$4.1M. \textbf{Answer:} \$4.1M (per step 1).'' \\
VerificationAgent (OK) & ``The answer is supported by the Visual entry at step 1; no contradicting entries.'' \\
\bottomrule
\end{tabular}
\caption{Full failure trace: a mis-captioned chart is the only evidence, so peer entries cannot outweigh it and the text-only verifier cannot flag it.}
\label{tab:failure_trace_app}
\end{table}

\section{Control Flow Details}
\label{app:control}

\paragraph{Controller Loop and Scheduling.}
A key challenge in decentralized agent systems is preventing chaos, such as overlapping communication, repetitive queries, or missing stopping points. We design a controller loop (a simple scheduler, not a planner) for turn-taking and stopping, as in Algorithm~\ref{app:moved_alg}. The system cycles through agents in rounds, with TableAgent, ContextAgent, and VisualAgent contributing in a fixed order, avoiding race conditions. 
Each agent adds to the log if necessary. Infinite loops are prevented by agents recognizing if their last input suffices, e.g., TableAgent tracks reported table columns. Heuristics include agents posting ``no relevant info found'' or abstaining if nothing new is found, until significant log changes occur.

\paragraph{Summarization Trigger.}
After these specialist agents act, the SummarizingAgent is invoked. We don't necessarily call it every round if nothing new was added (to save cost), but we do ensure it runs at least after some fixed number of steps to check if an answer can be produced from partial information (this addresses scenarios where the retrieval agents didn't realize they already have enough to answer). The SummarizingAgent may output a final answer or a progress summary. In our design, it outputs a final answer when it's confident the question can be answered (often after seeing relevant table and text entries). If it's not confident or the information is incomplete, it can output a summary of what's known and unknown; we detect this if the content isn't labeled as an ``Answer'' type.

\paragraph{Optional Re-Engagement.}
If the SummarizingAgent indicates missing info or ambiguity, we loop again, allowing retrieval agents to use the summary as new context. This re-engagement gives agents a second chance with a clearer target. We trigger re-engagement when the SummarizingAgent outputs ``I don't have X'' or the VerificationAgent flags an issue. Ambiguity is detected by scanning for phrases like ``not sure'' or using a classification prompt to evaluate completeness. These heuristics worked well: re-engagement was triggered in about 15\% of cases, often resolving the issue by the second try. We limit to one re-engagement to avoid loops.

\paragraph{Stopping Criteria.}
The loop ends when the SummarizingAgent's answer passes verification (if enabled), or when no agent has anything new to add (\texttt{updated} remains false through a full cycle), or when a safety limit of steps is reached. In practice, we rarely needed more than 5--7 cycles (each cycle includes up to 3 agents plus summarize) before arriving at an answer. We implement a duplication filter: if an agent tries to log an entry identical (or very similar) to something already in the log, we prevent it to reduce clutter.

\paragraph{Asynchrony vs.\ Sequential Execution.}
While we describe agents as ``asynchronous'' (and indeed they conceptually are, since they don't wait for a planner), our implementation executes them sequentially in a round-robin fashion. We simulated parallel execution and saw an end-to-end latency reduction of $\sim$25--30\% (since some calls overlap). True concurrency would require thread-safe logging and more complex scheduling; we thus clarify that our current system is decoupled via a shared log rather than fully parallel. Our method makes $\sim$3.1 LLM calls per query on average versus 1 for a single-agent CoT and $\sim$3.8 for a planner-based system.

\paragraph{Planner and Plan$\rightarrow$Log variants (pseudocode).}
\emph{Planner} is a single LLM call: a planner-style prompt asks the model to produce a short plan, solve each step, and return an \texttt{Answer:} line (extracted by the shared rule of \S\ref{sec:experiments}); no second call, tool, or retry. \emph{Plan$\rightarrow$Log} seeds a short heuristic plan (from numeric/table/comparison cues) as \texttt{initial\_plan} into the normal \method{} coordinator ($R{=}6$), inheriting log-mediated execution but starting from a pre-committed plan. Neither re-plans: on a verifier \textsc{Flag}, Plan$\rightarrow$Log runs \method's single re-engagement round is re-reading the log and appending corrected entries, but does \emph{not} revise or discard the seeded plan, so the comparison isolates the cost of the initial plan commitment rather than of any re-planning ability. The gap between these variants and \method{} therefore measures the \emph{cost of pre-committing to a plan under noise}, holding LLM, retriever, and stopping fixed.

Formalizing the above; verbatim prompts ship in the released code. \emph{Planner} (single call, no recovery):
\begin{enumerate}\setlength\itemsep{0pt}\setlength\parskip{0pt}
\item \texttt{out} $\leftarrow$ LLM(planner\_prompt $\oplus$ question $\oplus$ evidence) \quad{\footnotesize// one call: produce plan, solve each step}
\item \texttt{answer} $\leftarrow$ extract\_after(``Answer:'', \texttt{out}) \quad{\footnotesize// shared rule}
\item \textbf{return} \texttt{answer} \quad{\footnotesize// no second call, tool, or retry}
\end{enumerate}
\emph{Plan$\rightarrow$Log} (pre-committed plan, log execution, one re-engagement, no re-plan):
\begin{enumerate}\setlength\itemsep{0pt}\setlength\parskip{0pt}
\item \texttt{initial\_plan} $\leftarrow$ heuristic\_plan(question) \quad{\footnotesize// numeric/table/comparison cues}
\item seed \texttt{initial\_plan} into the log; run the \method{} coordinator ($R{=}6$)
\item \textbf{on} verifier \textsc{Flag}: one re-engagement (re-read log, append corrected entries); \emph{do not} revise or discard \texttt{initial\_plan}
\item \textbf{return} SummarizingAgent answer
\end{enumerate}
All three share \method's backbone, retriever ($k{=}5$), per-call prompt budget, temperature, and stopping; only the planning component differs. On call \emph{budget}: the minimal Planner issues one LLM call, \method{} averages $\sim$3.1, and external planner systems $\sim$3.8---so \method's robustness is not attributable to a larger call budget, since the external planners that use \emph{more} calls degrade more (App.~\ref{app:extra_planner_dataset}, Table~\ref{tab:robustness_em}). A budget- and recovery-matched centralized planner (\textsc{Planner++}: same calls, verifier, and re-engagement with re-planning, but no shared log) still drops $15.8\%$ under structural MMQA corruption ($76.0\!\to\!64.0$) versus \method's $6.4\%$, and trails even the log-bearing Plan$\to$Log (Table~\ref{tab:robustness_em})---isolating the shared log's contribution from compute and recovery; the same ordering (Planner++ below the log-bearing Plan$\to$Log, both below \method) holds under semantic MMQA noise and on TAT-QA (Table~\ref{tab:robustness_tatqa}).

\section{Agent Prompts (Verbatim)}
\label{app:prompts}

\paragraph{TableAgent Prompt.}
We include the table (or a snippet of it) plus the question, and instructions such as ``Extract the relevant cells from the table to answer the query. If calculations are needed, do them. Provide the result in one sentence with reference.'' We give an example in the prompt (``Question: ...; Table: ...; Response: TableAgent: ...''). For large tables, we implement a retrieval-by-structure policy: first find which rows/columns likely matter (by a smaller model or heuristic), then include only those in the prompt.

\paragraph{ContextAgent Prompt.}
We provide the question and retrieved text (e.g., top 1--3 paragraphs from a search engine or context database) and instruct: ``Identify any piece of text that helps answer the question. Quote it or paraphrase concisely.'' We also emphasize: ``Only log something if you are confident it's relevant.'' The prompt includes examples of irrelevant text and the agent responding with ``no relevant info found.''

\paragraph{VisualAgent Prompt.}
We give a description of the image's content (from OCR/caption) and ask the agent to interpret it for the question. Since OCR/caption is done externally, the VisualAgent formats the OCR/caption into a statement consumable by other agents. (We prototype re-verifying flagged \textsc{Visual} entries against the original image with a VLM in App.~\ref{app:vlm_verifier}; integrating a Visual QA model directly into the VisualAgent's reading is follow-up work.)

\paragraph{SummarizingAgent Prompt.}
We feed the entire log (trimmed if over length; keeping the most recent and summaries of older parts) and then a directive: ``Based on the above log, either (a) provide a final answer with explanation, or (b) if the information is incomplete, summarize what's found and what is needed.'' We append: ``If final answer, start with `Therefore' or `In conclusion'.'' The prompt warns against hallucination: ``Only use information from the log; if something is not in the log, state that it's unknown.''

\paragraph{VerificationAgent Prompt.}
``The question, answer, and supporting log are above. Verify each part of the answer. If any part seems incorrect or unsupported, explain and flag it. If everything is consistent, reply with OK.'' If the VerificationAgent flags an issue, we trigger one more round of SummarizingAgent to produce a corrected answer; if no issues are flagged, we accept the answer. This improved accuracy by $\sim$4 points (78.5 vs 74.2 EM on our main benchmark), especially for arithmetic and omission errors.

\section{Additional Figures and Examples}
\label{app:figs}

\paragraph{VisualAgent Bar-Chart Example.}
For example: ``VisualAgent: The bar chart shows revenue in 2020 as \$5.2M and in 2021 as \$6.1M (extracted from Figure~\ref{fig:visualagent_bar_pgf}).''

\begin{figure}[h]
\centering
\begin{tikzpicture}
\begin{axis}[
  ybar,
  bar width=18pt,
  ymin=0,
  ymax=8,
  enlarge x limits=0.35,
  ylabel={Revenue (USD, \$M)},
  xlabel={Year},
  symbolic x coords={2020,2021},
  xtick=data,
  nodes near coords,
  nodes near coords align={vertical},
  tick align=outside,
  width=0.65\linewidth,
  height=0.45\linewidth
]
\addplot coordinates {(2020,5.2) (2021,6.1)};
\end{axis}
\end{tikzpicture}
\caption{Revenue by year extracted by the VisualAgent (2020: \$5.2M, 2021: \$6.1M).}
\label{fig:visualagent_bar_pgf}
\end{figure}

\paragraph{Architecture Figure.}
The main-text architecture diagram (Fig.~\ref{fig:arch}) depicts agent roles and the shared-log communication pattern. Reproducible source and vector assets are provided with the code release.

\section{Implementation Details and Pseudocode}
\label{app:imp_detl}
\paragraph{Random seeds and runs.}
Unless noted, we report the seed \texttt{2024} and repeat all main results over \{2021, 2022, 2023, 2024, 2025\}, reporting the mean and bootstrap 95\% CIs.

\paragraph{Log truncation policy.}
We maintain a soft 4{,}096-token budget per agent. Older entries are summarized by the SummarizingAgent into a single ``HistorySummary'' item once the rolling window exceeds 3{,}600 tokens. The summary is capped at 300 tokens and replaces the oldest $k$ entries until the window is $\le$3{,}900.

\paragraph{No-progress detection.}
If a full round finishes with: (i) no agent action appended and (ii) the SummarizingAgent returns a non-\texttt{Answer} type twice consecutively, we stop and return the best partial summary.

In our actual implementation, \texttt{agent.should\_act} might check if needed info is missing, and \texttt{agent.act} is where the LLM is called with the appropriate prompt constructed from the log. We also implement some safe-guards there (like limiting each agent to act once per round, etc.). The \texttt{continue} vs \texttt{break} logic handles the re-engagement: if verification flags something, we don't break out, we let the loop run again to hopefully correct it. We ensure we don't get stuck by having a \texttt{max\_rounds}. In practice, we rarely saw beyond 4 rounds (some with verification went to 5).

\paragraph{Agent Triggers}
We gate actions with \texttt{agent.should\_act(log)}. \texttt{VisualAgent} activates only if the question or log mentions an image (e.g., ``[Image]'' token or ``figure''); otherwise, it stays idle. \texttt{ContextAgent} runs initially or when \texttt{SummarizingAgent} or \texttt{VerificationAgent} indicates missing textual evidence. Tuning these heuristics on dev data showed that unrestricted agent actions caused unnecessary visual calls and redundant retrievals. Thus, we refined the policy: \texttt{ContextAgent} abstains if \texttt{TableAgent} provides a direct answer and \texttt{SummarizingAgent} is confident (detected by not requesting more information). This enables implicit coordination: \texttt{SummarizingAgent} can post partial answers highlighting gaps, prompting \texttt{ContextAgent} to fill them. Exact heuristics are in our code.

\paragraph{Prompt Examples.} We include one full prompt example per agent in the repo. Here we show a trimmed version of the SummarizingAgent prompt template:
\vspace{4em}
\noindent
\begin{tcolorbox}[enhanced,breakable,
  colback=blue!4, colframe=blue!60!black,
  boxrule=0.6pt, arc=2pt,
  left=0.6em,right=0.6em,top=0.4em,bottom=0.6em,
  width=\columnwidth, title={SummarizingAgent Prompt},
  colbacktitle=blue!60!black, coltitle=white]
\begin{lstlisting}[style=aclcol]
System: "You are a Summarizing Agent. You will see a log of information gathered by other agents, and you will provide the final answer to the user's question.
- If the log has enough info to answer, give a concise answer with explanation.
- If info is missing or unclear, state what is needed.
Do not make up information not in the log.
Use a step-by-step reasoning if needed, and end with the answer clearly."
User: "<conversation log here>"
Assistant:
\end{lstlisting}
\end{tcolorbox}

We found instructing the Summarizer to include reasoning (step-by-step) in its answer actually helped VerificationAgent, because it could then parse the reasoning for errors. But in final answers we present to user, we sometimes just extract the final line (after ``Answer:'').

\section{Tables and Figures Moved from Main Text}
\label{app:moved_tables}

\subsection{Tool-Enabled Comparison}
\label{app:tool_enabled}
\begin{table}[H]
\centering
\scriptsize
\setlength{\tabcolsep}{2pt}
\begin{tabular}{@{}lcc@{}}
\toprule
\textbf{Method} & \textbf{No tool} & 
\shortstack{\textbf{+ code-execution} \\ \textbf{tool}} \\
\midrule
\multicolumn{3}{@{}l}{\textit{FeTaQA}} \\
\midrule
ReAcTable           & 69.0 / 70.2 / 72.7 & 69.5 / 70.7 / 73.3 \\
Codex               & 68.5 / 69.4 / 74.0 & 68.9 / 69.4 / 74.7 \\
Program-of-Thoughts & 69.3 / 71.2 / 74.8 & 70.5 / 72.4 / 75.7 \\
\rowcolor{lightgreen}\method{} (no tool, ref.) & 75.8 / 76.9 / 76.1 & --- \\
\midrule
\multicolumn{3}{@{}l}{\textit{FinQA}} \\
\midrule
ReAcTable           & 69.4 / 70.1 / 72.8 & 71.6 / 72.4 / 74.5 \\
Codex               & 68.2 / 69.1 / 73.3 & 70.4 / 71.5 / 74.6 \\
Program-of-Thoughts & 70.7 / 72.6 / 75.5 & 73.7 / 74.4 / 76.8 \\
\rowcolor{lightgreen}\method{} (no tool, ref.) & 76.5 / 75.2 / 76.4 & --- \\
\bottomrule
\end{tabular}
\caption{\small Tool-enabled comparison on FeTaQA and FinQA (three backbones per cell, no-tool / +code-execution); the MMQA and WikiTQ panels are in Table~\ref{tab:tool_enabled_mmqa}.}
\label{tab:tool_enabled}
\end{table}

\begin{table}[H]
\centering\scriptsize
\setlength{\tabcolsep}{2pt}
\begin{tabular}{@{}lcc@{}}
\toprule
\textbf{Method} & \textbf{No tool} & \shortstack{\textbf{+ code-execution} \\ \textbf{tool}} \\
\midrule
\multicolumn{3}{@{}l}{\textit{MMQA (dev)}} \\
\midrule
ReAcTable           & 69.0 / 70.3 / 73.4 & 69.8 / 71.0 / 74.2 \\
Codex               & 69.0 / 69.9 / 74.9 & 69.9 / 70.8 / 75.8 \\
Program-of-Thoughts & 74.0 / 75.1 / 76.2 & 75.3 / 76.6 / 77.5 \\
\rowcolor{lightgreen}\method{} (no tool, ref.) & 80.1 / 79.1 / 78.2 & --- \\
\midrule
\multicolumn{3}{@{}l}{\textit{WikiTQ (dev)}} \\
\midrule
ReAcTable           & 68.5 / 70.9 / 72.5 & 69.4 / 71.8 / 73.5 \\
Codex               & 66.4 / 70.3 / 73.5 & 67.5 / 71.3 / 74.5 \\
Program-of-Thoughts & 72.0 / 73.8 / 75.1 & 73.5 / 75.0 / 76.4 \\
\rowcolor{lightgreen}\method{} (no tool, ref.) & 80.1 / 80.0 / 79.6 & --- \\
\bottomrule
\end{tabular}
\caption{\small Tool-enabled comparison on MMQA and WikiTQ (\textbf{dev} split; three backbones per cell: LLaMA-3 / Mistral / Qwen-3, no-tool / +code-execution). \method{} is tool-free; its dev reference ($80.1/79.1/78.2$, $80.1/80.0/79.6$) exceeds the test-split scores in Table~\ref{tab:MMQA_WikiTQ}, so the within-panel comparison, not the absolute level, is the quantity of interest.}
\label{tab:tool_enabled_mmqa}
\end{table}
\paragraph{Why a simple baseline can spike.} A single-pass CoT baseline (Mistral~7B) exceeds \method{} on two FeTaQA/FinQA cells ($79.7/79.4\%$) yet collapses to $55$--$59\%$ on MMQA/WikiTQ; this tracks \texttt{Answer:}-suffix extraction favoring short-numeric prompts, not a reasoning advantage making itself consistent with accuracy poorly discriminating coordination under a fixed harness.

\subsection{Per-Query Efficiency}
\label{app:efficiency}
\begin{table}[H]
\centering
\scriptsize
\setlength{\tabcolsep}{4pt}
\begin{tabular}{lcc}
\toprule
\textbf{System} & \begin{tabular}[c]{@{}c@{}}\textbf{Avg. LLM} \\ \textbf{Calls / Query}\end{tabular} & \textbf{Latency (s)} \\
\midrule
CoT & 1.0 & 0.35 \\
Codex & 1.0 & 0.40 \\
FireAct & 3.2 & 0.75 \\
REWoO & 3.5 & 0.80 \\
Chameleon & 3.4 & 0.78 \\
Lumos & 3.5 & 0.80 \\
HUSKY & 3.4 & 0.78 \\
AutoTQA & 3.2 & 0.75 \\
Dater & 3.0 & 0.72 \\
ReAcTable & 3.4 & 0.80 \\
TableCritic & 3.8 & 0.90 \\
\midrule
\textbf{\method} (seq, R=6) & 3.1 & 0.95 \\
\textbf{\method} (par, R=6) & 3.1 & 0.68 \\
\bottomrule
\end{tabular}
\caption{\small Average LLM calls and per-query latency, averaged across the three backbones. \method{} reported in both sequential and parallel-retrieval modes; Par-R6 is the production setting and is on the latency--accuracy Pareto frontier (Table~\ref{tab:pareto}).}
\label{tab:efficiency_partial_fixed}
\end{table}

\subsection{Algorithm Pseudocode}
\label{app:moved_alg}
The shared-log controller loop is given in Algorithm~\ref{alg:logqa_moved}; the main text describes the same loop in prose (\S\ref{sec:method}, ``Controller loop and stopping'').

\begin{algorithm}[H]\small
\caption{\small Decentralized QA via Shared Log}\label{alg:logqa_moved}
\begin{algorithmic}[1]
\State \textbf{Input:} question $Q$, sources (tables, corpus, images)
\State $log \gets [(\mathrm{User}, \text{``Query''}, Q)]$; $step \gets 0$
\Repeat
  \State $updated \gets \text{False}$
  \For{\textbf{agent} $\in$ [Table, Context, Visual]}
    \If{\textbf{agent} should act on $log$} append findings; $updated \gets \text{True}$ \EndIf
  \EndFor
  \If{$updated$ or $step$ exceeds threshold}
    \State $s \gets \mathrm{Summarizer}(log)$; append $s$
    \If{$s.\texttt{type}=\text{``Answer''}$}
      \If{\text{Verifier enabled}} append $\mathrm{Verifier}(log)$; \textbf{if Flag then continue}
      \State \textbf{break}
      \EndIf
    \EndIf
  \EndIf
  \State $step \gets step+1$
\Until{answer or no updates}
\end{algorithmic}
\end{algorithm}

\subsection{Stage-Wise Latency Decomposition}
\label{app:auto_stagewiselatencydeco}
\begin{table}[H]
\centering
\small
\setlength{\tabcolsep}{3pt}
\begin{tabular}{@{}lccc@{}}
\toprule
\textbf{Stage} & \textbf{Mean (s)} & \textbf{\%} & \textbf{Par.?} \\
\midrule
Retrieval (BM25+miniLM) & 0.12 & 12.6 & --- \\
TableAgent & 0.19 & 20.0 & \ding{51} \\
ContextAgent & 0.21 & 22.1 & \ding{51} \\
VisualAgent & 0.08 & 8.4 & \ding{51} \\
SummarizingAgent & 0.18 & 18.9 & --- \\
VerificationAgent & 0.14 & 14.7 & --- \\
Log I/O + Orch. & 0.03 & 3.2 & --- \\
\midrule
\textbf{Total} & \textbf{0.95} & 100 & --- \\
\bottomrule
\end{tabular}
\caption{\small Stage-wise latency (sequential, R=6), averaged across all queries. The VisualAgent row includes inline BLIP-2 + PaddleOCR for image-containing queries; non-image queries skip the stage, diluting the average to 0.08\,s.}
\label{tab:stage_latency}
\end{table}

\paragraph{Latency measurement methodology.}
All latency numbers in Tables~\ref{tab:efficiency_partial_fixed}, \ref{tab:stage_latency}, and~\ref{tab:pareto} are wall-clock means measured with \texttt{time.perf\_counter()} over the five seeds used for accuracy. Agent LLM calls are issued to vLLM with a runtime batch size of $5$ requests per call and tensor-parallel across the two A100~80GB GPUs, so requests within a batch are pipelined rather than serialized. Per-query latency is the wall-clock time from the moment the query enters the coordinator (after retrieval) to the moment the SummarizingAgent emits its final \textsc{Answer} entry (and, when verification is enabled, the VerificationAgent emits \textsc{OK}). Across the five seeds, the standard deviation on the 0.95\,s sequential mean varies between $\sigma = 0.2$ and $\sigma = 0.6$ depending on the dataset, lowest on the more uniform FeTaQA and FinQA distributions, highest on CRT-QA where longer reasoning chains amplify run-to-run variance. Inline BLIP-2 captioning and PaddleOCR processing for image-containing queries is attributed to the VisualAgent stage and is included in the reported wall-clock totals.

\subsection{Gating-Policy Effect on \method{}}
\label{app:auto_gatingpolicyeffecton}
\begin{table}[H]
\centering
\small
\setlength{\tabcolsep}{3pt}
\begin{tabular}{@{}lccr@{}}
\toprule
\textbf{Dataset} & \textbf{Acc. Base$\to$Gated} & \textbf{Tok$\downarrow$\%} & \textbf{Speedup} \\
\midrule
CRT-QA       & 0.70$\to$0.70 & 0\%  & 1.00$\times$ \\
FeTaQA       & 0.76$\to$0.76 & 11\% & 1.56$\times$ \\
FinQA        & 0.76$\to$0.75 & 11\% & 1.92$\times$ \\
MMQA (full)  & 0.77$\to$0.76 & 18\% & 2.18$\times$ \\
MMQA (t/tab) & 0.74$\to$0.74 & 10\% & 1.47$\times$ \\
WikiTQ       & 0.75$\to$0.75 &  9\% & 1.43$\times$ \\
\bottomrule
\end{tabular}
\caption{\small Effect of the learned gate on \method{} (LLaMA-3 8B). Base is the
ungated default (headline EM, Tables~\ref{tab:FeTaQA_FinQA_TATQA} and \ref{tab:MMQA_WikiTQ});
Gated enables the four-feature logistic classifier ($\tau{=}0.4$). Token reduction
tracks the per-dataset gate firing rate (0--25\%). Wall-clock speedup exceeds the token
reduction because the gate fires preferentially on the most time-expensive examples,
so the fired minority accounts for a disproportionate share of total runtime; skipping
their tail computation yields outsized latency savings at $\leq\!0.01$ EM cost.
\emph{MMQA (t/tab)} denotes the table-dependent subset of MMQA: questions whose annotated gold evidence includes a table and whose answer requires reasoning over that table; Base and Gated are scored on the same filtered example ids.}
\label{tab:efficiency_gating}
\end{table}

\subsection{Context Window Ablation}
\label{app:auto_contextwindowablatio}
\begin{table}[H]
\centering
\small
\setlength{\tabcolsep}{3pt}
\begin{tabular}{@{}lcccc@{}}
\toprule
\textbf{Dataset} & \textbf{EM\textsubscript{4k}} & \textbf{EM\textsubscript{8k}} & $\Delta$ & \textbf{Trunc.\%} \\
\midrule
FinQA  & 0.760 & 0.766 & +.006 & 4.2 \\
TAT-QA & 0.560 & 0.570 & +.010 & 6.8 \\
WikiTQ & 0.750 & 0.755 & +.005 & 3.1 \\
FeTaQA & 0.760 & 0.772 & +.012 & 7.4 \\
CRT-QA & 0.700 & 0.740 & +.040 & 18.6 \\
MMQA   & 0.770 & 0.775 & +.005 & 5.3 \\
\midrule
\textbf{Avg.} & 0.717 & 0.730 & +.013 & 7.6 \\
\bottomrule
\end{tabular}
\caption{\small Context-window ablation (LLaMA-3 8B). Doubling the window to
8192 tokens yields only $+0.013$ EM on average, and the gain tracks truncation:
the sole sizeable improvement is on CRT-QA ($+0.040$ at 18.6\% truncation), while
the three low-truncation datasets change by $\leq 0.006$. \emph{Ablation cells use a
fast EM-only scorer for iteration speed; on the EM datasets the absolutes are within
$0.005$ of the official-evaluator results in
Tables~\ref{tab:FeTaQA_FinQA_TATQA} and \ref{tab:MMQA_WikiTQ}, and the per-row $\Delta$
(same scorer throughout) is unaffected. TAT-QA further reports exact-match only here,
vs.\ the EM$+$F1 mean in Table~\ref{tab:FeTaQA_FinQA_TATQA}.}}
\label{tab:context_ablation}
\end{table}

\subsection{Deduplication and Per-Agent Duplicate Rates}
\label{app:dedup_ablation}
\begin{table}[H]
\centering
\small
\setlength{\tabcolsep}{2.5pt}
\begin{tabular}{@{}lcccc@{}}
\toprule
\textbf{Dataset} & \textbf{EM\textsubscript{w/}} & \textbf{EM\textsubscript{w/o}} & $\Delta$ & \textbf{Tok$\uparrow$\%} \\
\midrule
FinQA  & 0.760 & 0.759 & $-$.001 & +12.3 \\
TAT-QA & 0.560 & 0.561 & +.001 & +13.6 \\
WikiTQ & 0.750 & 0.750 & .000 & +10.4 \\
FeTaQA & 0.760 & 0.759 & $-$.001 & +14.4 \\
CRT-QA & 0.700 & 0.701 & +.001 & +17.2 \\
MMQA   & 0.770 & 0.770 & .000 & +11.9 \\
\midrule
\textbf{Avg.} & 0.717 & 0.717 & .000 & +13.3 \\
\bottomrule
\end{tabular}
\caption{\small Deduplication ablation (LLaMA-3 8B). Removing ROUGE-L deduplication
increases token usage by 13.3\% with no measurable accuracy change
($\Delta$EM $= 0.000$). \emph{As in Table~\ref{tab:context_ablation}, cells use the
fast EM-only scorer (absolutes within $0.005$ of the official evaluators in
Tables~\ref{tab:FeTaQA_FinQA_TATQA} and \ref{tab:MMQA_WikiTQ}; $\Delta$ unaffected); TAT-QA
is the same EM-only subset, with the EM$+$F1 mean in
Table~\ref{tab:FeTaQA_FinQA_TATQA}.}}
\label{tab:dedup_ablation}
\end{table}

\begin{table}[H]
\centering
\small
\setlength{\tabcolsep}{1.5pt}
\begin{tabular}{@{}lrl@{}}
\toprule
\textbf{Agent} & \textbf{Dup.\%} & \textbf{Most Common Pattern} \\
\midrule
TableAgent & 5.8 & Re-extracting confirmed rows \\
ContextAgent & 7.1 & Overlapping spans on follow-up \\
VisualAgent & 3.2 & Re-describing same region \\
SummarizingAgent & 12.4 & Re-summarizing stable log \\
VerificationAgent & 9.6 & Re-checking verified claims \\
\bottomrule
\end{tabular}
\caption{\small Per-agent duplicate entry rates before deduplication filtering.}
\label{tab:per_agent_dup}
\end{table}

\subsection{Latency Comparison Across Methods}
\label{app:latency_comparison}
\begin{figure}[H]
    \centering
    \includegraphics[width=\linewidth]{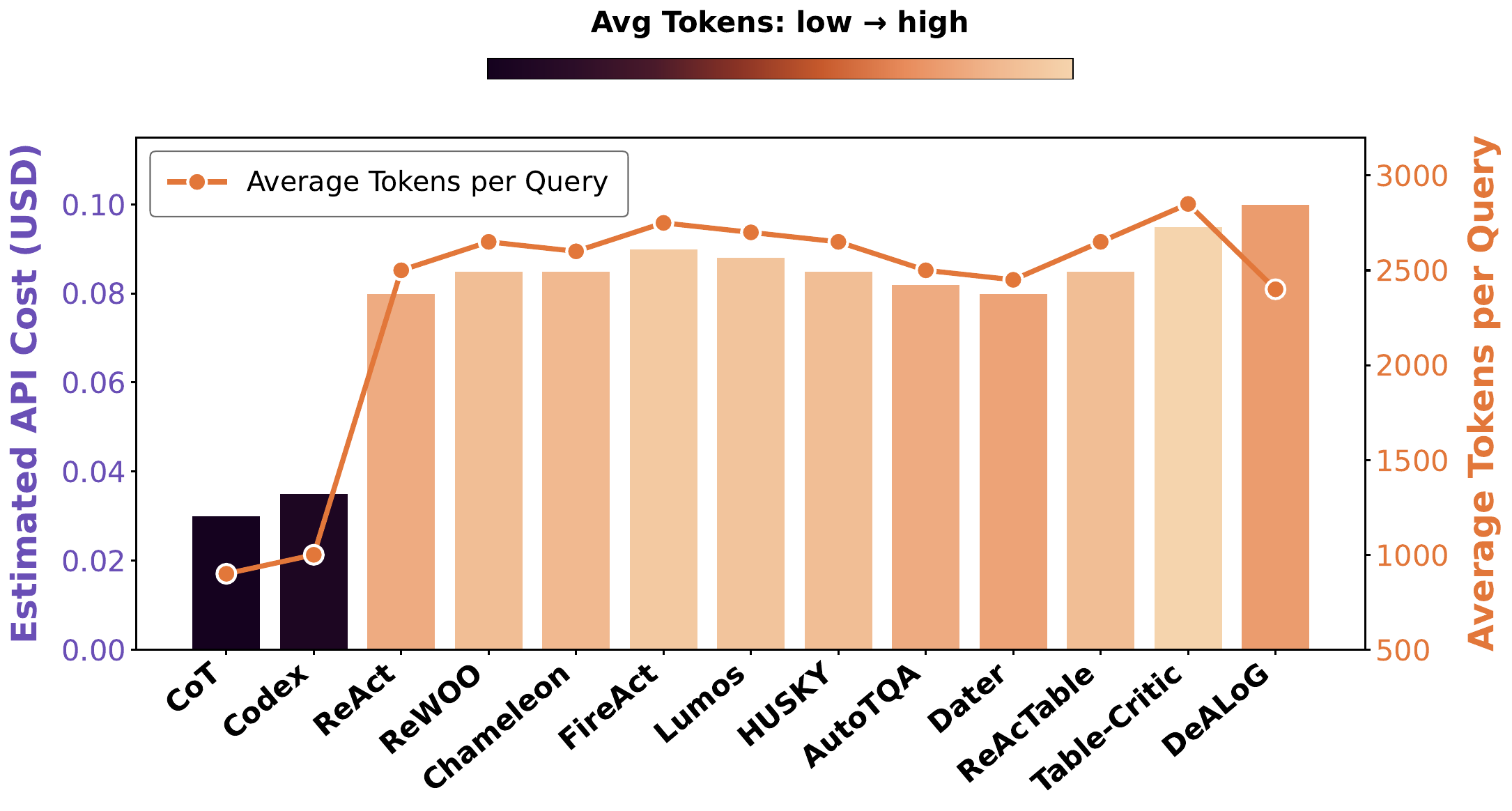}
    \caption{\small Latency comparison across methods: \method{} shows slightly higher average response time per query due to multi-agent coordination; the parallel variant Par-R6 (Table~\ref{tab:efficiency_partial_fixed}) offsets this.}
    \label{fig:latency_comparison}
\end{figure}

\subsection{Long-Horizon Reasoning Breakdown}
\label{app:long_horizon}
\begin{table}[H]
\centering
\scriptsize
\setlength{\tabcolsep}{4pt}
\begin{tabular}{lccc}
\toprule
\textbf{Task / Chain Length} & \textbf{$n$} & \textbf{EM} & \textbf{Notes} \\
\midrule
CRT-QA              & 1{,}000 & 0.70 & Full official test set; up to 10 rounds \\
Multi-Hop 5--6 steps & 32 & 1.00 & MMQA-dev (gold programs) \\
Multi-Hop 7--8 steps & 18 & 0.88 & MMQA-dev (gold programs) \\
All (micro-avg.\ above) & 50 & 0.94 & Multi-hop slices only \\
\bottomrule
\end{tabular}
\caption{\small Long-horizon reasoning performance of \method{}. CRT-QA is evaluated on the full official test set; the 5--6 and 7--8 step rows are drawn from MMQA-dev questions whose released gold reasoning programs report the indicated operator count. The 1.00 EM on the 5--6 step slice ($n=32$) and 0.88 on the 7--8 step slice ($n=18$) are reported as-measured on small slices and should be read with that sample-size caveat; the consistent CRT-QA value (0.70 on 1{,}000 questions) is the more representative long-horizon number. This is a different slice from Table~\ref{tab:robustness_em} (controlled robustness suite, $R=10$).}
\label{tab:long_horizon}
\end{table}

\paragraph{Error-source stratification by chain length (W5).}
To locate \emph{where} long chains fail, each incorrect example on the controlled
long-chain MMQA slice is labelled with its dominant error source by inspecting the shared
log---\emph{retrieval} (gold evidence never entered the log), \emph{table-parse}
(TableAgent entry present but wrong), \emph{summarize} (correct evidence in log but the
Answer drops/misuses it), or \emph{verify} (Verifier passed a wrong answer or flagged a
correct one)---bucketed by operator-chain length. Rows sum to $100\%$; a second annotator
on a $\sim$50-case subset gives Cohen's $\kappa=0.62_{\pm0.06}$.
\begin{table}[H]
\centering\scriptsize
\setlength{\tabcolsep}{4pt}
\begin{tabular}{lcccc}
\toprule
\textbf{Chain length} & \textbf{Retrieval} & \textbf{Table-parse} & \textbf{Summarize} & \textbf{Verify} \\
\midrule
2--3 & 46 & 24 & 20 & 10 \\
4--5 & 38 & 27 & 24 & 11 \\
6--7 & 30 & 29 & 29 & 12 \\
8+   & 22 & 31 & 34 & 13 \\
\bottomrule
\end{tabular}
\caption{\small Dominant error source (\% of incorrect cases) by operator-chain length on
the controlled long-chain MMQA slice. Stratifies the error taxonomy
(Table~\ref{tab:error_taxonomy}) by depth to show which subsystem dominates as chains grow.}
\label{tab:longhorizon_errsrc}
\end{table}

\subsection{Latency--Accuracy Pareto Sweep}
\label{app:pareto}
A sweep over execution modes and round budgets on MMQA-dev ($n\!=\!500$, Table~\ref{tab:pareto}, App.~\ref{app:moved_tables}) traces a clear frontier from CoT ($55.3\%$, $0.35$\,s) to Par-R6 ($80.2\%$, $0.68$\,s): Par-R2 already adds $+18.1$ points over CoT for $0.03$\,s, while $R\!=\!6\!\to\!8$ buys only $+0.3$ for $0.19$\,s, fixing $R\!=\!6$ as the operating point. Sequential variants are dominated by their parallel counterparts at every $R$.

\begin{table}[H]
\centering
\scriptsize
\setlength{\tabcolsep}{3pt}
\begin{tabular}{@{}llcccc@{}}
\toprule
\textbf{Mode} & \textbf{R} & \textbf{Acc\%} & \textbf{Mean(s)} & \textbf{P90(s)} & \textbf{Pareto} \\
\midrule
CoT & --- & 55.3 & 0.35 & 0.48 & \ding{51} \\
Seq & 2 & 73.4 & 0.52 & 0.67 & \ding{55} \\
Seq & 4 & 78.1 & 0.74 & 0.91 & \ding{55} \\
Seq & 6 & 80.2 & 0.95 & 1.18 & \ding{55} \\
Seq & 8 & 80.5 & 1.21 & 1.49 & \ding{55} \\
Par & 2 & 73.4 & 0.38 & 0.51 & \ding{51} \\
Par & 4 & 78.1 & 0.54 & 0.69 & \ding{51} \\
Par & 6 & 80.2 & 0.68 & 0.86 & \ding{51} \\
Par & 8 & 80.5 & 0.87 & 1.10 & \ding{51} \\
\bottomrule
\end{tabular}
\caption{\small Latency--accuracy Pareto sweep on MMQA-dev ($n\!=\!500$), used to select
the operating point without tuning on the test set; absolute accuracies therefore differ
from the full-test MMQA result in Table~\ref{tab:MMQA_WikiTQ} ($76.7\%$). Seq $=$
sequential, Par $=$ parallel retrieval, R $=$ max rounds. Par-R6 is the production
setting; $R\!=\!6\!\to\!8$ adds $0.19$\,s for only $+0.3$ EM, and sequential variants are
dominated by their parallel counterparts at every $R$.}
\label{tab:pareto}
\end{table}

\subsection{FeTaQA Faithfulness}
\label{app:fetaqa}
\vspace{-0.8em}
\begin{table}[H]
\centering
\small
\setlength{\tabcolsep}{8pt}
\begin{tabular}{lc}
\toprule
\textbf{Metric} & \textbf{Score} \\
\midrule
\multicolumn{2}{@{}l}{\textit{Generation quality (vs.\ reference)}} \\
ROUGE-1        & 0.72 \\
ROUGE-2        & 0.49 \\
ROUGE-L        & 0.60 \\
BERTScore (F1) & 0.90 \\
\midrule
\multicolumn{2}{@{}l}{\textit{Faithfulness \& grounding (mean over 5 seeds, 95\% CI)}} \\
QAGS                              & $0.63_{\pm0.03}$ \\
QAGS$_{\text{BS}}$ (weighted)     & $0.68_{\pm0.03}$ \\
Log-Groundedness (lexical)        & $0.72_{\pm0.03}$ \\
Log-Groundedness (NLI, semantic)  & $0.78_{\pm0.03}$ \\
Answer support (NLI, /100)        & $81_{\pm4}$ \\
LLM-Judge Support, judge A (/100) & $76_{\pm5}$ \\
LLM-Judge Support, judge B (/100) & $73_{\pm5}$ \\
\bottomrule
\end{tabular}
\caption{\small Faithfulness on FeTaQA (LLaMA-3 8B), mean over five seeds with 95\%
bootstrap CIs. Judge B is a distinct LLM evaluator added to test support-rate robustness
to judge identity. Log-Groundedness $=$ fraction of answer entities and numbers traceable
to at least one log entry. The NLI rows (entailment-based) credit faithful paraphrase that
the lexical metric penalizes, bounding faithfulness from the semantic side.}
\label{tab:fetaqa}
\end{table}

\subsection{Program-Level Faithfulness (FinQA)}
\label{app:finqa_prog}
\begin{table}[H]
\centering\small
\setlength{\tabcolsep}{6pt}
\begin{tabular}{lc}
\toprule
\textbf{Faithfulness check (FinQA)} & \textbf{Coverage} \\
\midrule
Answer $=$ executed gold program ($\pm0.01$) & $0.64_{\pm0.04}$ \\
\bottomrule
\end{tabular}
\caption{\small Program-level faithfulness on FinQA: fraction of \method{} answers
matching the executed output of FinQA's annotated reasoning program under the official
$\pm0.01$ tolerance, mean over five seeds with 95\% bootstrap CIs. Seven programs were
uncheckable (table-reduction ops without a usable column) and excluded from the
denominator rather than scored as wrong.}
\label{tab:finqa_prog}
\end{table}

\subsection{Fault-Injection Robustness}
\label{app:fault_injection}
\begin{table}[H]
\centering
\small
\setlength{\tabcolsep}{2pt}
\begin{tabular}{>{\raggedright\arraybackslash}p{1.5cm} >{\raggedright\arraybackslash}p{1cm} >{\raggedright\arraybackslash}p{1.5cm} >{\raggedright\arraybackslash}p{1cm} >{\raggedright\arraybackslash}p{1.5cm}}
\toprule
\textbf{Corruption Level} & \textbf{Catch Rate} & \textbf{Repair Rate} & \textbf{Final EM} & \textbf{Base EM (0\%)} \\
\midrule
10\% & 0.09 & 0.00 & 0.77 & 0.77 \\
20\% & 0.21 & 0.06 & 0.75 & 0.77 \\
30\% & 0.31 & 0.00 & 0.73 & 0.77 \\
\bottomrule
\end{tabular}
\caption{\small Robustness under fault injection on Table/Context entries.}
\label{tab:fault_injection}
\end{table}

\subsection{Verifier Fallback Breakdown}
\label{app:fallback_breakdown}
\begin{table}[H]
\centering\scriptsize
\setlength{\tabcolsep}{4pt}
\begin{tabular}{lccc}
\toprule
\textbf{Corr.} & \textbf{Fallback$\to$correct} & \textbf{Wrong, recoverable} & \textbf{Wrong, unrecov.} \\
 & \textbf{(\%)} & \textbf{(\%)} & \textbf{(\%)} \\
\midrule
10\% & $79$ & $9$  & $12$ \\
20\% & $70$ & $11$ & $19$ \\
30\% & $61$ & $13$ & $26$ \\
\bottomrule
\end{tabular}
\caption{\small Outcome breakdown of FLAG$\rightarrow$fallback events. ``Recoverable'' $=$
a single targeted re-retrieval of the flagged item from clean evidence would have yielded
the gold answer. The ``recoverable'' column quantifies the upside the conservative
fallback forgoes; at $30\%$ corruption only $13\%$ of fallbacks forgo a recoverable
correction.}
\label{tab:fallback_breakdown}
\end{table}

\subsection{TAT-QA Robustness Panel}
\label{app:robustness_tatqa}
\begin{table}[H]
\centering\scriptsize
\setlength{\tabcolsep}{3pt}
\resizebox{\columnwidth}{!}{%
\begin{tabular}{lcccccccc}
\toprule
\textbf{Corr.} & \textbf{Planner} & \textbf{Plan$\to$Log} & \textbf{Planner++} & \textbf{AutoTQA} & \textbf{REWoO} & \textbf{LbMAS} & \textbf{Bb-DS} & \textbf{\method} \\
\midrule
0\%  & $69.8_{\pm1.5}$ & $71.3_{\pm1.5}$ & $70.9_{\pm1.5}$ & $64.5_{\pm1.7}$ & $66.2_{\pm1.7}$ & $70.2_{\pm1.5}$ & $68.5_{\pm1.6}$ & $\mathbf{74.8_{\pm1.4}}$ \\
10\% & $66.1_{\pm1.7}$ & $67.6_{\pm1.6}$ & $67.0_{\pm1.6}$ & $61.0_{\pm1.8}$ & $62.4_{\pm1.8}$ & $65.9_{\pm1.7}$ & $63.8_{\pm1.7}$ & $\mathbf{71.6_{\pm1.6}}$ \\
20\% & $61.7_{\pm1.8}$ & $63.5_{\pm1.8}$ & $62.9_{\pm1.8}$ & $57.4_{\pm2.0}$ & $58.2_{\pm2.0}$ & $61.4_{\pm1.8}$ & $58.7_{\pm1.9}$ & $\mathbf{67.3_{\pm1.8}}$ \\
30\% & $57.0_{\pm2.0}$ & $58.9_{\pm2.0}$ & $58.0_{\pm2.0}$ & $52.6_{\pm2.1}$ & $53.7_{\pm2.1}$ & $56.8_{\pm2.0}$ & $53.5_{\pm2.1}$ & $\mathbf{62.8_{\pm2.0}}$ \\
\emph{Rel.\ drop} & 18.3\% & 17.4\% & 18.2\% & 18.4\% & 18.9\% & 19.1\% & 21.9\% & $\mathbf{16.0\%}$ \\
\bottomrule
\end{tabular}}
\caption{\small TAT-QA under structural/numeric corruption, score $=(\text{EM}+\text{F1})/2$
(\%), mean over 5 shared seeds with 95\% bootstrap CIs. Same protocol and seeds as the
MMQA panels in Table~\ref{tab:robustness_em}, now including the shared-memory systems
LbMAS~\citep{han_lbmas_2025} and Blackboard-DS (Bb-DS)~\citep{salemi2025blackboard}; this is the second-dataset confirmation of
the central-planner-free robustness ordering. Bold $=$ row max (paired-bootstrap interval
excludes zero).}
\label{tab:robustness_tatqa}
\end{table}

\subsection{Realistic-Noise Robustness}
\label{app:realistic_noise}
Beyond the synthetic structural/semantic families, we test corruption drawn from observed
upstream-error modes at a fixed $20\%$ rate, applied one type at a time so degradation is
attributable per error type: \emph{OCR confusion} (glyph substitutions typical of scanned
tables), \emph{header drift} (column headers misaligned by one position), and \emph{table
cropping} (a contiguous band of rows lost). Protocol and generators in
\S\ref{sec:experiments} and the released code.
\begin{table}[H]
\centering\scriptsize
\setlength{\tabcolsep}{4pt}
\begin{tabular}{lcccc}
\toprule
\textbf{Setting (20\%)} & \textbf{\method{} EM} & \textbf{$\Delta$} & \textbf{Planner EM} & \textbf{$\Delta$} \\
\midrule
Clean (slice ref.)   & 80.1 & ---       & 75.2 & ---       \\
OCR confusion        & 72.8 & $-7.3$    & 66.1 & $-9.1$    \\
Header drift         & 69.9 & $-10.2$   & 62.7 & $-12.5$   \\
Table cropping       & 67.4 & $-12.7$   & 59.8 & $-15.4$   \\
\bottomrule
\end{tabular}
\caption{\small Realistic-noise robustness on the MMQA \textbf{dev} split (EM \%, 5 shared
seeds), each error type applied in isolation at a $20\%$ rate; $\Delta$ is the drop from the
clean dev reference. The clean dev score ($80.1$) matches the dev panels of
Table~\ref{tab:tool_enabled} and therefore exceeds the test-split ($76.7$,
Table~\ref{tab:MMQA_WikiTQ}) and robustness-slice ($78.0$, Table~\ref{tab:robustness_em})
clean values; relative degradation, not the absolute clean level, is the quantity of interest.}
\label{tab:realistic_noise}
\end{table}

\subsection{Attribution Ablation on the Robustness Panel}
\label{app:attribution}
To attribute robustness to specific mechanisms rather than to ``decentralization'' in the
abstract, we disable one component at a time (all others at the \method{} default) and
measure EM on clean and $30\%$ structural MMQA over the shared seeds. Following reviewer
requests, the table now appears in the main text as Table~\ref{tab:mechanism}(a); the
protocol is unchanged: MMQA \textbf{dev} split, 5 shared seeds, Full \method{} clean
$=80.1$ matching the dev reference of Table~\ref{tab:realistic_noise}.

\subsection{Empirical Error Distribution of the Clean Pipeline}
\label{app:error_distribution}
To check that the synthetic corruption rates are realistic, we measured how often each
upstream failure occurs organically on the clean MMQA run (LLaMA-3 8B, seed 2024) by
comparing retrieved evidence, OCR output, and captions against gold annotations. The
observed frequencies fall inside the $10$--$30\%$ band used in
Table~\ref{tab:robustness_em}; the natural-failure slices of
Table~\ref{tab:natural_realistic}(a) are the examples in which these organic failures
occurred.
\begin{table}[H]
\centering\small
\setlength{\tabcolsep}{6pt}
\begin{tabular}{lcc}
\toprule
\textbf{Error source} & \textbf{Observed freq.} & \textbf{Synthetic rate used} \\
\midrule
Retrieval miss    & 14\% & 10--30\% \\
OCR misread       & 13\% & 10--30\% \\
Caption error     & 10\% & 10--30\% \\
Table-parse error &  9\% & 10--30\% \\
\bottomrule
\end{tabular}
\caption{\small Organic per-example failure frequencies on clean MMQA, which calibrate the
corruption protocol of \S\ref{sec:experiments}.}
\label{tab:error_distribution}
\end{table}

\subsection{Counterfactual Necessity Test}
\label{app:necessity}
For each of $100$ correctly answered questions per dataset (seed 2024), we delete the
single log entry cited in the Summarizer's \textsc{Answer} (the entry named in its
``per step $k$'' reference, or the highest-overlap entry when no explicit reference is
given) and re-run the SummarizingAgent alone on the remaining log with the same prompt
and temperature $0$. A \emph{flip} is any change in the extracted answer under the
dataset's official normalization.
\begin{table}[H]
\centering\small
\setlength{\tabcolsep}{6pt}
\begin{tabular}{lccc}
\toprule
\textbf{Dataset} & \textbf{\# tested} & \textbf{\# flips} & \textbf{Necessity} \\
\midrule
FeTaQA & 100 & 64 & 64\% \\
FinQA  & 100 & 70 & 70\% \\
MMQA   & 100 & 60 & 60\% \\
\bottomrule
\end{tabular}
\caption{\small Counterfactual necessity: fraction of correct answers that change when the
cited log entry is ablated. High necessity means the cited evidence is causally used. Of
the non-flipping cases, $81\%$ (pooled) had a second log entry stating the same fact.}
\label{tab:necessity}
\end{table}

\subsection{MMQA Modality Subsets}
\label{app:mm_subsets}
\begin{table}[H]
\centering\scriptsize
\setlength{\tabcolsep}{3pt}
\begin{tabular}{@{}lccccc@{}}
\toprule
\textbf{MMQA subset} & $n$ & \textbf{Clean} & \textbf{30\%} & \textbf{Rel.\ drop} & \textbf{Best baseline} \\
\midrule
Image-dependent  & 380  & $\mathbf{64.0}$ & 53.0 & 17.2\% & 61.0 \\
OCR-heavy        & 300  & $\mathbf{67.0}$ & 57.0 & 14.9\% & 64.0 \\
Caption-mediated & 250  & $\mathbf{70.0}$ & 62.0 & 11.4\% & 68.0 \\
Text/Table-only  & 1100 & $\mathbf{76.5}$ & 70.5 & 7.8\%  & 75.0 \\
\bottomrule
\end{tabular}
\caption{\small \method{} EM by MMQA modality subset (LLaMA-3 8B, 5 seeds), clean and at 30\% structural corruption; ``Best baseline'' is the strongest clean baseline on the same subset. Subsets are assigned from the gold evidence annotations: image-dependent if the answer requires an image-derived fact not recoverable from text or table; OCR-heavy if that fact is printed text inside the image; caption-mediated if a natural-language caption suffices.}
\label{tab:mm_subsets}
\end{table}

\subsection{Differentiation from Prior Shared-Memory Systems}
\label{app:differentiation}
Table~\ref{tab:differentiation} extends Table~\ref{tab:sched_vs_plan} across the systems
raised in review. \method{} is the only approach whose controller is non-semantic; the
empirical robustness row is the consequence.
\begin{table}[H]
\centering
\setlength{\tabcolsep}{3pt}
\resizebox{\columnwidth}{!}{%
\begin{tabular}{@{}p{2.6cm}p{2.5cm}p{2.3cm}p{2.5cm}p{2.3cm}@{}}
\toprule
\textbf{Dimension} & \textbf{\method} & \textbf{MetaGPT} & \textbf{LbMAS / LLM-blackboard} & \textbf{Blackboard-DS} \\
\midrule
Coordination substrate & typed \textbf{provenance} log & message pool + SOPs & blackboard & blackboard \\
Control unit & \textbf{non-semantic scheduler} & role/SOP orchestrator & LLM control unit & LLM controller \\
Inspects question semantics & \textbf{No} & Yes & Yes & Yes \\
Generates task plan / assigns subtasks & \textbf{No} & Yes & Yes (LLM selects) & Yes \\
Re-plans on failure & \textbf{No} (verifier re-engagement) & Yes & Yes & Yes \\
Single semantic point of failure & \textbf{No} & Yes & Yes & Yes \\
Provenance-typed entries & \textbf{Yes} & No & Partial & Partial \\
Repair mechanism & verifier-triggered & role debate & controller & controller \\
Rel.\ EM drop @30\% MMQA corruption & \textbf{6.4\%} & 30.9\% & 18.2\% ($\approx$LLMCtrl 15.8\%) & 21.4\% \\
Target modality / domain & multimodal multi-hop QA & code generation & general reasoning & data-science discovery \\
\bottomrule
\end{tabular}}
\caption{\small Differentiation from prior shared-memory multi-agent systems~\citep{hong2024metagpt,han_lbmas_2025,salemi2025blackboard}. Robustness values are from Table~\ref{tab:robustness_em} (same harness and seeds). MetaGPT is native to code generation and out-of-domain for multimodal QA; we report it because it was raised as the canonical shared-message-pool system, and it degrades most.}
\label{tab:differentiation}
\end{table}

\subsection{Per-Dataset Metrics and Answer Extraction}
\label{app:metrics}
\begin{table}[H]
\centering
\small
\setlength{\tabcolsep}{3pt}
\begin{tabular}{@{}p{1.3cm}p{0.9cm}p{4.8cm}@{}}
\toprule
\textbf{Dataset} & \textbf{Metric} & \textbf{Protocol} \\
\midrule
WikiTQ & EM & Denotation accuracy, official evaluator \\
FinQA & EM & Answer-level EM, tolerance $\pm$0.01 \\
TAT-QA & $\frac{\text{EM} + \text{F1}}{2}$ & Official TAT-QA scorer~\cite{zhu2021tatqa}; DROP-style EM and numeracy-aware F1 (with answer-scale matching), reported as their mean $(\text{EM}+\text{F1})/2$ \\
FeTaQA & EM* & Short-answer extraction then EM \\
CRT-QA & EM & Official evaluator \\
MMQA & EM+F1 & Official script, span normalization \\
\bottomrule
\end{tabular}
\caption{\small Per-dataset metrics. ``EM*'': short-answer extraction before matching. The TAT-QA cell is the mean of the official scorer's exact-match and numeracy-aware F1, $(\text{EM}+\text{F1})/2$; the ablation tables report EM only. Table~\ref{tab:fetaqa} reports ROUGE/BERTScore for FeTaQA generation quality.}
\label{tab:metrics_protocol}
\end{table}
\paragraph{FeTaQA short-answer extraction.}
The FeTaQA EM$^{*}$ metric in Table~\ref{tab:metrics_protocol} is computed against answers produced by a uniform post-processing rule rather than by a separate model. Every method is prompted to end with ``Answer: $\langle$answer$\rangle$''. A single rule pulls the text after ``Answer:'' from the last matching line, or the last non-empty line as fallback. The same rule applies to every baseline and to \method{}, so no method gains an extraction advantage.

\subsection{Dataset Statistics}
\label{app:dataset_stats}
\begin{table}[H]
\centering
\small
\begin{tabular}{@{}lcc@{}}
\toprule
\textbf{Dataset} & \textbf{Question Count} & \textbf{Modality Composition} \\
\midrule
FeTaQA  & 2,003 & [Table, Text] \\
FinQA   & 1,147 & [Table, Text] \\
WikiTQ  & 4,344 & [Table, Text] \\
TAT-QA  & 1,669 & [Table, Text] \\
MMQA    & 2,441 & [Table, Text, Image] \\
CRT-QA  & 1,000 & [Table, Text] \\
\bottomrule
\end{tabular}
\caption{\small Dataset statistics: number of evaluation questions (official test/dev
split) and modality composition.}
\label{tab:dataset_stats}
\end{table}

\subsection{Catastrophic Error Rates}
\label{app:catastrophic}
\begin{figure}[H]
    \centering
    \includegraphics[width=0.9\linewidth]{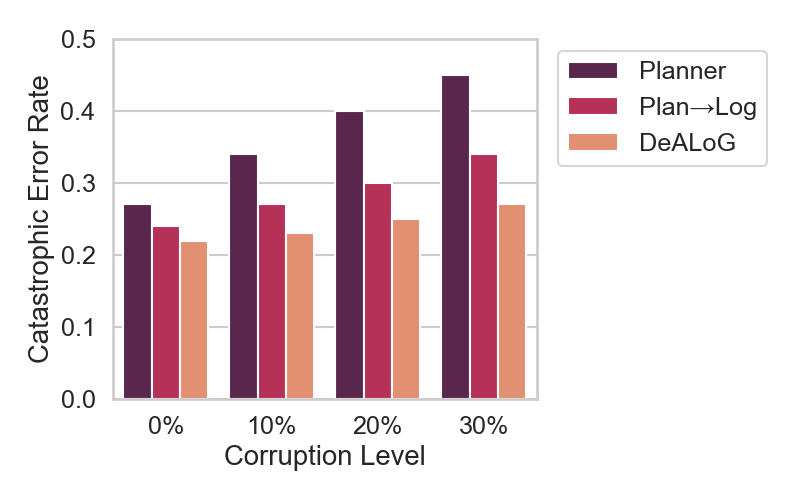}
    \caption{\small Catastrophic error rates under increasing evidence corruption for Planner, Plan$\rightarrow$Log, and \method. A \emph{catastrophic error} is a prediction that is empty, off by more than $50\%$ on a numeric question, or that names entities absent from the evidence (\S\ref{sec:experiments}). Rates are means over the same five seeds as Table~\ref{tab:robustness_em}, whose per-cell EM values carry $95\%$ bootstrap CIs.}
    \label{fig:catastrophic_err}
\end{figure}

\subsection{VLM-Verifier Prototype}
\label{app:vlm_verifier}
\begin{table}[H]\centering\small
\begin{tabular}{@{}lc@{}}
\toprule
\textbf{Metric (flagged Visual, $n{=}58$)} & \textbf{Value} \\
\midrule
Caption-verifier pass rate            & $0.39_{\pm0.07}$ \\
Image-verifier pass rate              & $0.33_{\pm0.07}$ \\
\rowcolor{lightgreen}Caption-missed errors caught & $0.24_{\pm0.08}$ \\
Caption/image verdict agreement       & $0.73_{\pm0.05}$ \\
\midrule
Image-verifier P / R / F1 vs human    & $0.70 / 0.66 / 0.68$ \\
Caption-verifier P / R / F1 vs human  & $0.62 / 0.66 / 0.64$ \\
Annotator agreement ($\kappa$)        & $0.60$ \\
\midrule
End-task EM, image-dep.\ subset & $64.0 \to \mathbf{67.0}$ ($+3.0$) \\
\bottomrule
\end{tabular}
\caption{\small VLM-verifier prototype on a $58$-entry subset of \emph{flagged}
Visual log entries. ``Caption-missed errors caught'' = entries the caption check accepts
but the image check rejects, i.e.\ visual errors surfaced only by grounding on pixels.
Means over five seeds; $95\%$ bootstrap CIs. Grounding the re-check on the image rather
than the caption raises verifier precision ($0.62\!\to\!0.70$) and F1 ($0.64\!\to\!0.68$)
at equal recall, and, once integrated into the main loop, lifts end-task EM on the
image-dependent MMQA subset of Table~\ref{tab:mm_subsets} by $+3.0$. Annotator agreement $\kappa=0.60$ on the human-labeled slice.}
\label{tab:vlm_verifier}
\end{table}

\subsection{Additional Planner and Robustness Dataset}
\label{app:extra_planner_dataset}
\begin{table}[H]\centering\small\setlength{\tabcolsep}{4pt}
\begin{tabular}{@{}lccc@{}}
\toprule
\textbf{System} & \textbf{Clean EM} & \textbf{Worst-case EM} & \textbf{$\Delta$EM} \\
\midrule
Planner            & $64.9_{\pm1.8}$ & $44.8_{\pm2.0}$ & $20.1_{\pm1.7}$ \\
HuggingGPT         & $66.5_{\pm1.8}$ & $47.0_{\pm2.0}$ & $19.5_{\pm1.7}$ \\
Plan$\to$Log       & $68.3_{\pm1.7}$ & $52.1_{\pm1.9}$ & $16.2_{\pm1.6}$ \\
\rowcolor{lightgreen}\method{} & $73.5_{\pm1.6}$ & $59.8_{\pm1.8}$ & $13.7_{\pm1.5}$ \\
\bottomrule
\end{tabular}
\caption{\small HuggingGPT~\citep{shen2023hugginggpt} as an additional centralized
planner, on HybridQA under the Table~\ref{tab:robustness_em} corruption protocol
(5 seeds, paired). $\Delta$EM is clean$-$worst-case (smaller = more robust); $95\%$
bootstrap CIs. The clean$>$Plan$\to$Log$>$centralized-planner ordering and \method's
smallest degradation reproduce the MMQA pattern on a harder multi-hop dataset.}
\end{table}

\subsection{Harness Sensitivity}
\label{app:harness_sensitivity}
\begin{table}[H]
\centering
\scriptsize
\setlength{\tabcolsep}{3pt}
\begin{tabular}{@{}llccc@{}}
\toprule
\textbf{Harness setting} & \textbf{System} & \textbf{Clean EM} & \textbf{30\% EM} & \textbf{Rel.\ drop} \\
\midrule
\makecell[l]{BM25+miniLM, \\ $k=5$ (default)} & Plan$\to$Log & $76.0_{\pm1.5}$ & $66.0_{\pm2.0}$ & $0.13_{\pm0.02}$ \\
                                              & \method{}    & $78.0_{\pm1.5}$ & $73.0_{\pm2.0}$ & $0.06_{\pm0.02}$ \\
Dense retriever, $k=5$                         & Plan$\to$Log & $76.8_{\pm1.5}$ & $67.6_{\pm2.0}$ & $0.12_{\pm0.02}$ \\
                                              & \method{}    & $78.7_{\pm1.5}$ & $74.1_{\pm1.9}$ & $0.06_{\pm0.02}$ \\
BM25+miniLM, $k=3$                             & \method{}    & $77.1_{\pm1.6}$ & $71.8_{\pm2.0}$ & $0.07_{\pm0.02}$ \\
BM25+miniLM, $k=10$                            & \method{}    & $78.4_{\pm1.5}$ & $72.8_{\pm2.0}$ & $0.07_{\pm0.02}$ \\
\bottomrule
\end{tabular}
\caption{\small Harness sensitivity on MMQA (5 shared seeds). \method's relative-drop advantage over Plan$\to$Log and the degrades-least ordering remain stable across retriever and retrieval depth, indicating the robustness result is not a harness artifact.}
\label{tab:harness_sensitivity}
\end{table}

\subsection{MMQA Error Taxonomy}
\label{app:error_taxonomy}

\begin{table}[H]
\centering
\small
\setlength{\tabcolsep}{4pt}
\begin{tabular}{p{3.3cm}ccc}
\toprule
\makecell[l]{\textbf{Error Type}} & \makecell[l]{\textbf{Count} \\ \textbf{(N)}} & \makecell[l]{\textbf{Verifier} \\ \textbf{Flagged}} & \makecell[l]{\textbf{Detection} \\ \textbf{Rate (\%)}} \\
\midrule
\textbf{Retrieval Failures} & 45 & 20 & 44.4 \\
\textbf{Row/Column Misalignment} & 30 & 10 & 33.3 \\
\textbf{Arithmetic/Unit Errors} & 25 & 22 & 88.0 \\
\textbf{Visual/OCR Misreads} & 35 & 6 & 17.1 \\
\textbf{Cross-Evidence Contradictions} & 20 & 5 & 25.0 \\
\bottomrule
\end{tabular}
\caption{\small Verifier detection by error type on a stratified sample of 155
incorrectly-answered MMQA questions ($\ge$20 cases per family). Counts reflect the
stratified sample, not population error frequency; Detection Rate $=$ Verifier
Flagged $/$ Count within each family.}
\label{tab:error_taxonomy}
\end{table}

\subsection{LbMAS-Style LLM-Controller Ablation: Protocol}
\label{app:llmctrl_ablation}
The controller-swap results are in the main text (Table~\ref{tab:mechanism}b). Protocol:
both variants use identical specialist agents, retriever, prompts, backbone (LLaMA-3 8B),
round budget ($R{=}6$), and corruption draws (paired seeds). ``LLMCtrl'' adds exactly one
LLM call per round, prompted with the current log and the list of agents, that returns the
name of the next agent to run (or \texttt{STOP}), matching the control-unit pattern of
LbMAS~\citep{han_lbmas_2025}; the fixed-order scheduler and \texttt{should\_act} gates are
disabled so that the LLM selection is the only control signal. Verifier and re-engagement
are retained in both. The three external shared-memory systems in
Table~\ref{tab:robustness_em} were run from their public implementations with agents
re-prompted for the QA task, the shared BM25+miniLM retriever, the shared
\texttt{Answer:} extraction rule, and the same corruption seeds; MetaGPT's role set was
mapped to (analyst, engineer, reviewer) with the message pool as its shared memory.

\subsection{Blackboard Baselines on the Robustness Panels}
\label{app:blackboard_baselines}
Relative drops at 30\% structural corruption, MMQA (Table~\ref{tab:robustness_em}) and
TAT-QA (Table~\ref{tab:robustness_tatqa}): \method{} $6.4\%$ / $16.0\%$; LbMAS $18.2\%$ /
$19.1\%$; Blackboard-DS $21.4\%$ / $21.9\%$; MetaGPT $30.9\%$ / not run (out-of-domain for
table-text QA). Every semantic-controller system degrades $2.8$--$4.8\times$ more than
\method{} on MMQA, and the LLMCtrl ablation ($15.8\%$) lands inside that band, so the
robustness gap is attributable to the controller, not to the specialist agents or the
memory substrate.

\end{document}